\documentclass[runningheads]{llncs}
\pdfoutput=1
\usepackage[pdftex]{graphicx}
\usepackage{wrapfig}
\usepackage{amsfonts}       
\usepackage{nicefrac}       
\usepackage{microtype}      
\usepackage{amsmath}
\usepackage{graphicx}
\usepackage{subfig}
\usepackage{caption}
\usepackage{inputenc}
\usepackage{booktabs}
\graphicspath{{figs/}}
\titlerunning{ML based training of infinite mixtures for uncertainty quantification}
\usepackage[dvipsnames]{xcolor}

\DeclareMathOperator{\KL}{D_{KL}}

\begin{document}
\title{
Investigating maximum likelihood based training of infinite mixtures for uncertainty quantification\thanks{Funded by the Deutsche Forschungsgemeinschaft (DFG, German Research Foundation) under Germany's Excellence Strategy - EXC 2092 \textsc{CaSa} - 390781972.}}
%
%
\author{Sina D\"aubener\orcidID{0000-0003-4057-9566} \and
Asja Fischer\orcidID{0000-0002-1916-7033}}
\authorrunning{S. D\"aubener and A. Fischer}
%
\institute{Ruhr University Bochum \\
Universitätsstraße 150, 44801 Bochum \\
\email{\{sina.daeubener, asja.fischer\}@rub.de}}
\maketitle              
\begin{abstract}
Uncertainty quantification in neural networks gained a lot of attention in the past years.
The most popular approaches, 
Bayesian neural networks (BNNs), Monte Carlo dropout, and deep ensembles have one thing in common: they are all based 
on some kind of mixture model. While the BNNs build infinite mixture models and are derived via variational inference, the latter two build finite 
mixtures trained with the maximum likelihood method. In this work we investigate the effect of training an infinite mixture distribution with the maximum likelihood method instead of variational inference. We find that 
the proposed objective
 leads to stochastic networks
 with an increased predictive variance, which improves uncertainty based identification of miss-classification and 
 robustness against adversarial attacks 
 in comparison to a standard BNN with equivalent network structure.
 The new model also displays higher entropy on
 out-of-distribution data.

\keywords{Uncertainty  \and Infinite Mixtures \and Robustness.}
\end{abstract}

\section{Introduction}
\label{sec:intro}
The thriving force behind uncertainty estimation is the hope to gain better model understanding and derive more reliable predictions. In recent years many methods capable of capturing uncertainty have been proposed, e.g. deep ensembles~\cite{lakshminarayanan2017simple}, Monte Carlo dropout~\cite{gal2016dropout}, and various kinds of Bayesian neural networks~\cite{swag-maddox,neal1995bayesian,louizos2017multiplicative,louizos_structured_2016,rank1dusenberry2020efficient}. These network types have been applied on various tasks such as out-of-distribution detection or defense against adversarial attacks. All of them can be seen as some kind of mixture model, either finite mixtures trained by maximizing the likelihood of their parameters or infinite mixtures trained by variational inference or Monte Carlo based approaches.

In this paper, we analyze infinite mixture distributions trained by a maximum likelihood method. We start by describing the approach and the analyzed uncertainty measures in section~\ref{sec:approach}, 
where we also formulate three hypotheses about the behavior of 
infinite mixtures trained with our novel objective compared to those trained with  variational inference 
with respect to 
their capability of uncertainty quantification. In section~\ref{sec:results} we conducted multiple experiments on the well known MNIST and Fashion MNIST data sets, which strengthen our formulated hypotheses. Last, we discuss possible future research directions based on our findings in section~\ref{sec:discussion}.

\section{Background and Approach}
\label{sec:approach}
Our approach is inspired by the 
observation  that  the most famous existing approaches for uncertainty quantification---namely
\textit{Bayesian neural networks} (BNNs), \textit{Monte Carlo (MC) dropout}, and \textit{Deep Ensembles (DEs)}---are based on 
some kind of mixture distribution.

In the following, 
let $x$ be some arbitrary input, $y$ the corresponding output, and $D = \{(x_1, y_1), ..., (x_N, y_N) \}$ a training data set consisting out of N samples.
The predictive distribution of MC dropout and DEs is given by a finite mixture of the form
\begin{equation}\label{eq:predictive_distribution}
    p(y|x) = \sum_{\theta} p(y|x,\theta)q(\theta) \enspace,
\end{equation}
where in the case of a DE of $k$ networks the mixing distribution is given by $q(\theta)=1/k ;~\forall \theta$, and where $q(\theta)$ specifies the probability of all dropout masks in the case of MC dropout. 

\textit{Bayesian neural networks} (BNNs), on the other hand, 
can be interpreted as an infinite mixture, 
\begin{equation}\label{eq:predictive_distribution}
    p(y|x) = \int_{\Theta} p(y|x,\theta)q(\theta)d\theta \enspace,
\end{equation}
where $q(\theta)$ 
represents an approximate posterior distribution over the parameters, and the integral is approximated via Monte Carlo sampling by a finite sum in practice.

One way to obtain the approximate posterior of BNNs is via variational inference (VI). The idea here is to fit the variational distribution $q(\theta)$ 
to approximate the true but untractable posterior $p(\theta|D)$. This is done by maximizing the so called \textit{evidence lower bound} (ELBO), which for 
a prior distribution  $p(\theta)$, is 
given by
\begin{equation}
        \mathcal{L}_{VI}= \sum_{n=1}^N \mathbb{E}_{q(\theta)}[\log p(y_n| x_n, \theta)]- \lambda \KL[q(\theta) || p(\theta)]
        \label{eq:VI-objective}
        \enspace,  \\
\end{equation}
where, $\KL[\cdot||\cdot]$ denotes the Kullback-Leibler divergence (KLD).
Note that for exact VI $\lambda = 1$. However, in practice $\lambda$ is often tuned for deriving better accuracy, which leads to a \textit{cold posterior} \cite{wenzel2020howgoodposterior}.
Further, the KLD in eq.~\eqref{eq:VI-objective} can be interpreted as playing the role of a penalization term, to keep the deviation of the approximate posterior from the prior small. 
For specific choices of $q(\theta)$ and $p(\theta)$ the KLD can be calculated exactly, while the expectation in the first term is approximated by drawing samples from  $q(\theta)$.

Contrary, DEs and MC dropout build on the maximum likelihood (ML) principle, where the log probability 
of the parameters
is maximized given 
the data set. While for a  DE all networks are trained by maximizing the likelihood of their parameters separately\footnote{ See \cite{rank1dusenberry2020efficient} for a discussion about parallel optimization. }, MC dropout networks are  
trained by maximizing the following objective
\begin{align}
    \mathcal{L}_{ML*}&= \sum_{n=1}^N \log \mathbb{E}_{q(\theta)}[p(y_n| x_n, \theta)] \hspace{4pt} \enspace,
\end{align}
where in practice 
the expectation 
is 
approximated based on a single sample from the dropout distribution $q(\theta)$.

In this paper we 
investigate the idea of making use of
a continuous distribution over neural network weights to define an infinite mixture model as done for BNNs but directly maximizing eq.~\eqref{eq:predictive_distribution} for a given set of  training samples 
by maximizing the likelihood of the parameters of $q(\theta)$.
This results in a \textit{maximum likelihood trained infinite mixture}. We further suggest to incorporate the KLD as a regularization term into the objective to keep the weights close to some prior as done by Kristiadi et. al \cite{cdn_wir}.

Hence, in our approach we will train 
an infinite mixture by maximizing the following objective
\begin{align}
    \mathcal{L}_{ML}&= \sum_{n=1}^N \log \mathbb{E}_{q(\theta)}[p(y_n| x_n, \theta)] - \lambda \KL [q(\theta) || p(\theta)] \label{eq:ML-objective} \enspace,
\end{align}
to which we will refer to as the ML objective in the rest of the paper.
Note, that the empirical approximation of $\mathcal{L}_{VI}$  (eq.~\eqref{eq:VI-objective})
gets equivalent to the empirical approximation of
$ \mathcal{L}_{ML}$  (eq.~\eqref{eq:ML-objective}) if the expectations are approximated with only one sample.

\subsection{Measuring uncertainty in mixture models}
\label{sec:uncertainty}

The  entropy of the predictive distribution over the output classes 
is often used as a measure of the prediction uncertainty 
of a  classification model. For a fixed input $x$ this entropy is given by 
\begin{equation}
    \mathcal{H}[p(y|x)] = -\sum_{c=1}^{K}  p(y_c|x) \cdot \log p(y_c|x) \enspace,
\end{equation}
where 
$K$ corresponds to the number of possible output classes.

In a mixture model, 
where the final prediction is defined as the expected prediction under the mixing distribution, the uncertainty of a prediction can alternatively be quantified by measuring the predictive variance of mixture components 
for multiple draws 
$\theta_s \sim q(\theta)$
from the 
mixing distribution, that is by estimating
\begin{equation}\label{eq:entropy}
\frac{1}{S}\sum_{s=1}^S p(y|x, \theta_s)^2 - p(y|x)^2 \enspace,
\end{equation}
where $p(y|x)$ is approximated by $\frac{1}{S}\sum_{s=1}^S p(y|x, \theta_s)$.

\subsection{Three hypotheses about the variance
behavior}
\label{subsec:hypo}

Next, we look at the two objectives in more detail to hypothesize about the potential differences they induce w.r.t.~the  variance between the predictions of different mixture components.
\begin{enumerate}
\item  
We first note, that obviously for both objectives 
(i) a prior over network parameters with a larger variance encourages the 
variance of the mixing distribution to be larger which we hypothesize  
might in turn increase  the final predictive variance  between the mixture components, while a prior with a smaller variance should induce the reverse effect, (ii) the impact of the prior should be stronger the higher the value of $\lambda$ is.

\item 
Let $(x_n,y_n)$ be an arbitrary but fixed data sample. Because of the Jensen inequality we know that the expectation term of the ML 
objective is bigger than that of the VI 
objective, that is
\begin{align}
\label{eq:jensen}
    \log \mathbb{E}_{q(\theta)}[p(y_n| x_n, \theta)]  \geq & \mathbb{E}_{q(\theta)}[\log p(y_n| x_n, \theta)]  \enspace .
\end{align}
Hence, if the KLD term is weighted equally in both objectives (in the sense that the same value for $\lambda$ is chosen), it has a larger weight relative to the expectation term in the VI objective. Therefore, for fixed $\lambda$ the choice of the prior should have more impact on the variance of the 
mixing distribution (i.e., the posterior) in the case of the VI based model.

\item Generally, the exact expectation is intractable, and thus is approximated by Monte Carlo sampling during training. This transforms eq.~\eqref{eq:jensen}
in the following way \begin{align}
       \log \left( \frac{1}{S} \sum_{s=1}^S p(y_n| x_n, \theta_s)  \right) \geq &  \frac{1}{S} \sum_{s=1}^S \log p(y_n| x_n, \theta_s) \label{eq:log_sum}\\
       = & \frac{1}{S}  \log \left( \prod_{s=1}^S p(y_n| x_n, \theta_s) \right)
       \enspace,
       \label{eq:log_prod}
\end{align}
where $S$ is again the number of the Monte Carlo samples drawn from $q(\theta)$.
By inspecting eq.~\eqref{eq:log_prod} we see 
that the expression gets large if 
the probabilities $p(y_n| x_n, \theta_s)$ 
are high for all $\theta_s$. In contrast a small value of $p(y_n| x_n, \theta_s)$ for a single sample $ \theta_s$ has a lower impact on the value of the expression on the left hand side of eq.~\eqref{eq:log_sum}.
 Therefore we hypothesize that the variance between the predictions of the mixture components 
 may tend to be smaller for a BNN than for an infinite mixture with equivalent model architecture trained with the ML objective. 
 \end{enumerate}

\section{Experiments}

\label{sec:results}
In this experimental analysis we aim to investigate the differences between an infinite mixture trained by the proposed ML objective and an infinite mixture formed by a BNN which was trained by VI\footnote{ An empirical comparison of infinite mixtures to finite mixtures based on a deep ensemble and MC dropout can be found section \ref{sec:infvsfin} in the appendix.}, and to test the validity of our hypotheses. Computationally, the difference in the training objectives reduces to an exchange of the expectation and logarithm in eq.~\eqref{eq:VI-objective}. As noted before $\mathcal{L}_{VI}$ and $\mathcal{L}_{ML}$  get equivalent, if the expectations are approximated with only one Monte Carlo sample. To leverage the difference we therefore always used $5$ samples for approximating the expectation.

As a base model we employ the BNN proposed by Louizos and Welling \cite{louizos_structured_2016}, who model the posterior distribution as a matrix variate normal distribution. 
If not stated otherwise, we chose a matrix variate normal distribution $\mathcal{MVN}(0,I,I)$ as the prior $p(\theta)$, which equates to a multivariate standard normal distribution. For modeling $p(y|x,\theta)$ we use a neural network with two hidden layers of size 128 each and 10 output neurons representing the classes in a softmax layer.
For our infinite mixture we employ an equivalent model architecture, i.e., we choose the mixing distribution $q(\theta)$ to be a matrix variate normal distribution, define $p(\theta)$ 
as a $\mathcal{MVN}(0,I,I)$, and use the same network to model $p(y|x,\theta)$,
but use $\mathcal{L}_{ML}$ instead of $\mathcal{L}_{VI}$ as training objective\footnote{Note, that the analysis can be easily extended to different network architectures such as those defined for mean field variational inference which has its origins in physics~\cite{MFVI} or multiplicative normalizing flows~\cite{louizos2017multiplicative}.}.

We conducted all our experiments on the well known MNIST and Fashion MNIST (FMNIST) data sets, which consists of black and white images of size $28\times 28$, labeled with one out of 10 possible classes.  We used  
the full training set of $60,000$ samples for training and the whole test set consisting out of $10,000$ samples at test time. When experimenting with adversarial examples those were estimated based on  the first $1,000$ test samples, if not stated otherwise.

During optimization we used ADAM \cite{kingma2014adam} with a mini batch size of 200 and an initial learning rate of $0.001$ for $30,000$ iterations (which corresponds to 100 epochs).
At test time all predictions are made based on $100$ samples drawn from $q(\theta)$ to approximate the integral of the predictive 
distribution in eq.~\eqref{eq:predictive_distribution}.
Each experimental setting was repeated 10 times with 10 different random seeds. 
We report means and standard deviations over the 10 trials in the tables and indicate the 3-fold standard deviations for better visibility  in our plots by error bars.

\subsection{The impact of the prior variance}
\label{subsec:priors}

We start our analysis by investigating the influence of the variance of the prior on (i) the  variance of the final mixing distribution and (ii) the predictive variance of mixture components for multiple draws from the mixing distribution, where we calculate the variance for each class separately and take the maximum over the ten classes for reporting the predictive variance.

\begin{table}[htb!]
\caption{Accuracy on test set for different prior variances.}
\centering
\resizebox{\textwidth}{!}{%
\begin{tabular}{l l |l l l l  } 
\toprule
Prior variance  &  & 0.5  & 1 & 1.5 & 3  \\ 
\midrule 
MNIST &  VI  & \textbf{0.975} $\pm 0.57\text{e-}3 \enspace $& 
0.974 $\pm 0.78 \text{e-}3 \enspace$ & 
0.973 $\pm 0.91 \text{e-}3 \enspace$ & 
0.970 $\pm 1.30 \text{e-}3 \enspace$ \\
&ML \ &\textbf{0.974} $\pm 0.83 \text{e-}3 \enspace $& 
0.973 $\pm 0.97 \text{e-}3 \enspace $& 
0.971 $\pm 0.78 \text{e-}3 \enspace $& 
0.969 $\pm 1.18 \text{e-}3 \enspace  $\\
\hline
FMNIST \ &VI & \textbf{0.857} $\pm 2.34\text{e-}3 \enspace$ & 
0.854 $\pm 2.28\text{e-}3 \enspace $& 
0.853 $\pm 2.33\text{e-}3 \enspace $& 
0.849 $\pm 2.82 \text{e-}3 \enspace$ \\
&ML & \textbf{0.864} $\pm 1.85\text{e-}3 \enspace$ & 
0.860 $\pm 2.53 \text{e-}3 \enspace$ & 
0.858 $\pm 1.62 \text{e-}3 \enspace$ & 
0.853 $\pm 2.32 \text{e-}3 \enspace$ \\
\bottomrule
\end{tabular}%
}
\label{tab:accuracy_priors}
\end{table}

\begin{figure}[bh!]\label{fig}
    \centering
    \subfloat[ML, MNIST]{\includegraphics[width=0.48\linewidth]{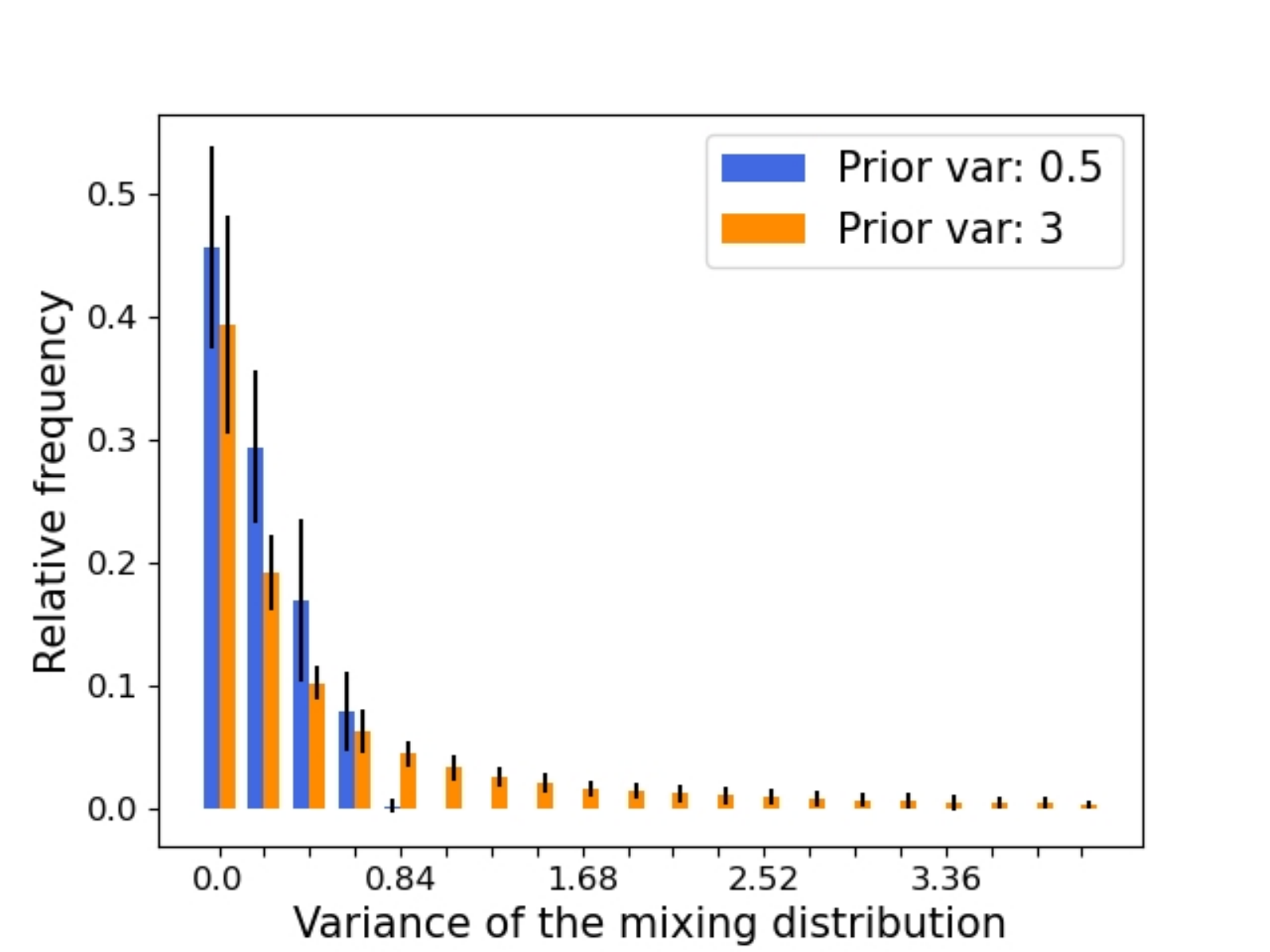}}
    \subfloat[VI, MNIST]{\includegraphics[width=0.48\linewidth]{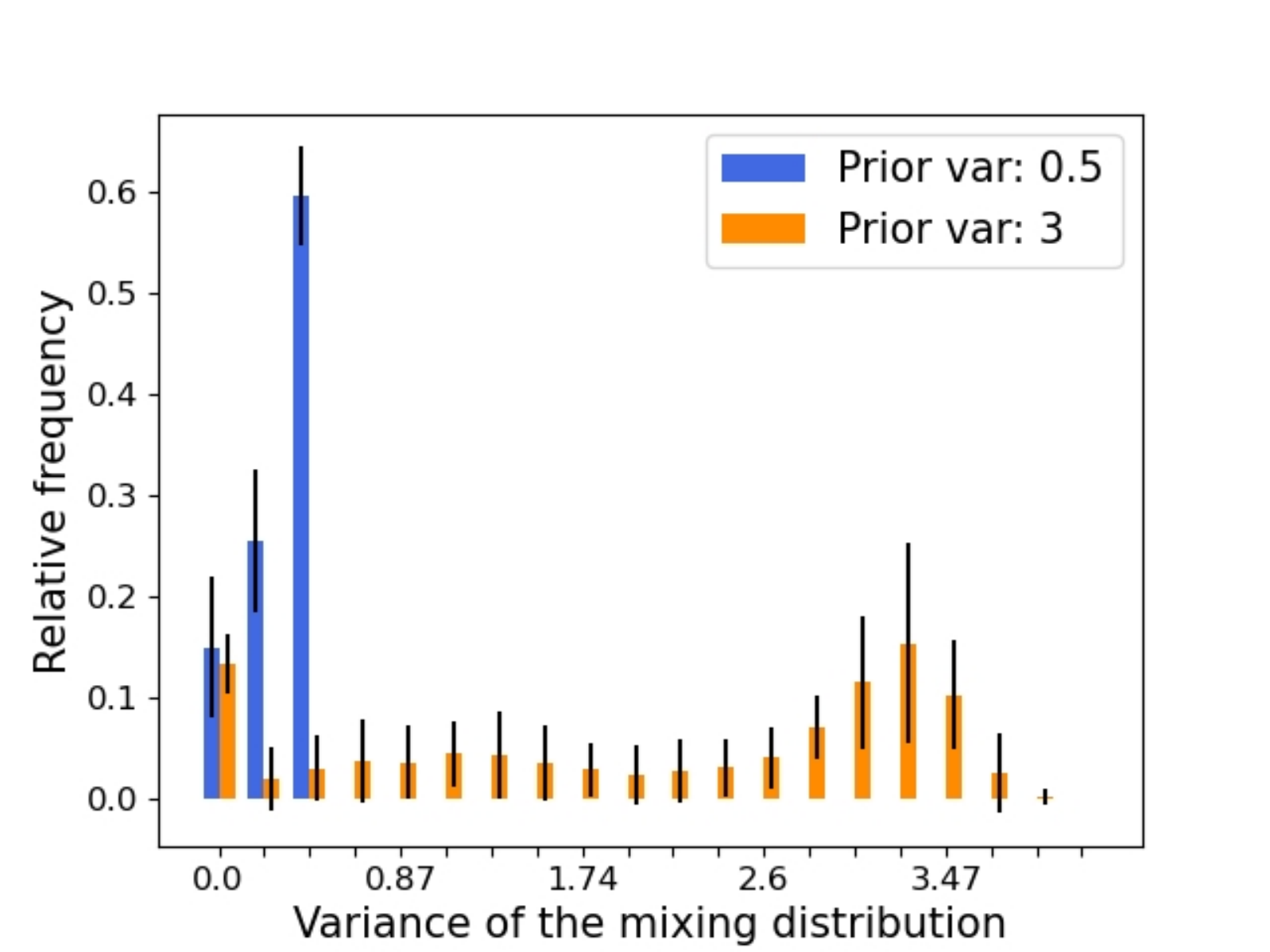}}
    \qquad
    \subfloat[ML, MNIST]{\includegraphics[width=0.48\linewidth]{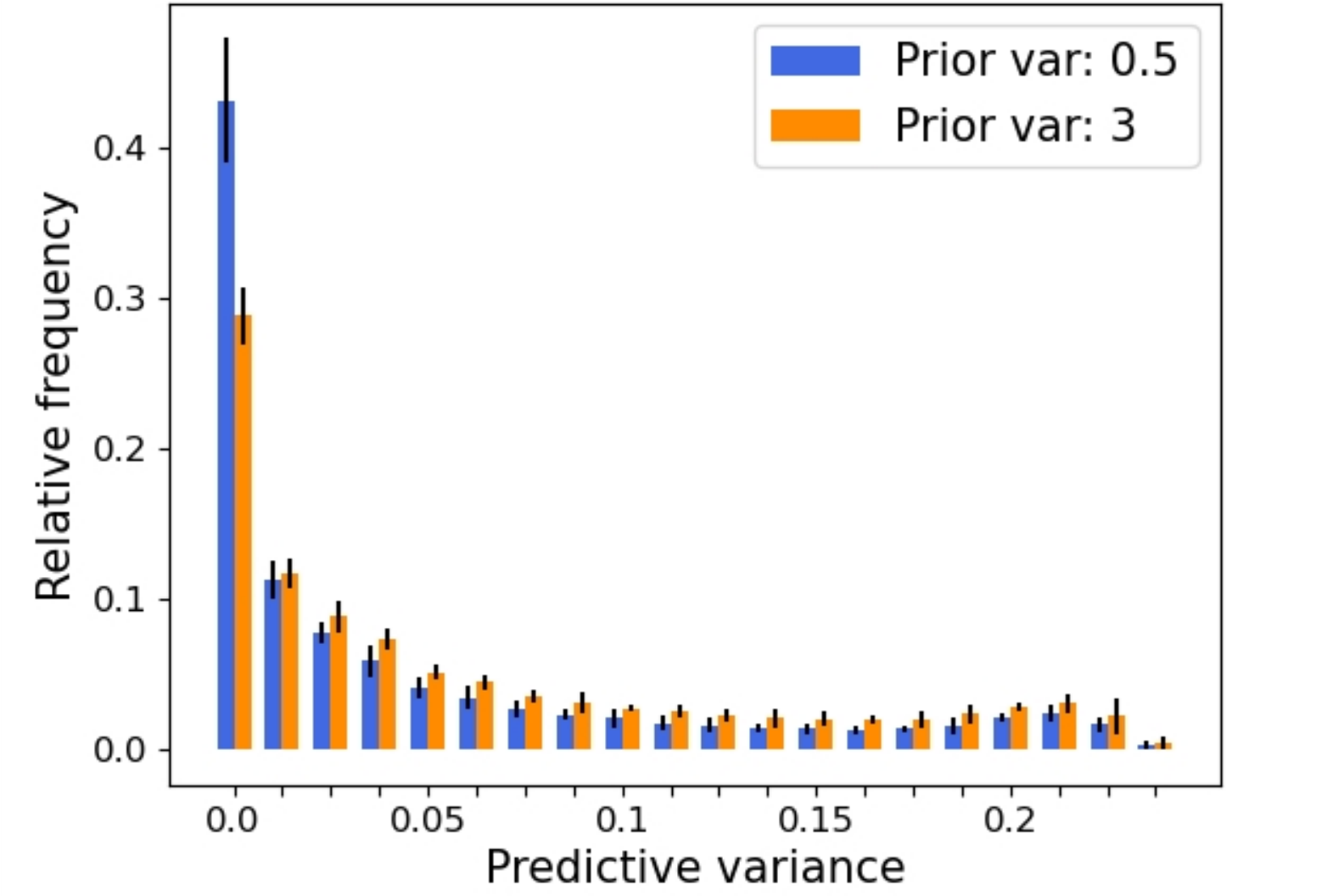}}
    \subfloat[VI, MNIST]{\includegraphics[width=0.48\linewidth]{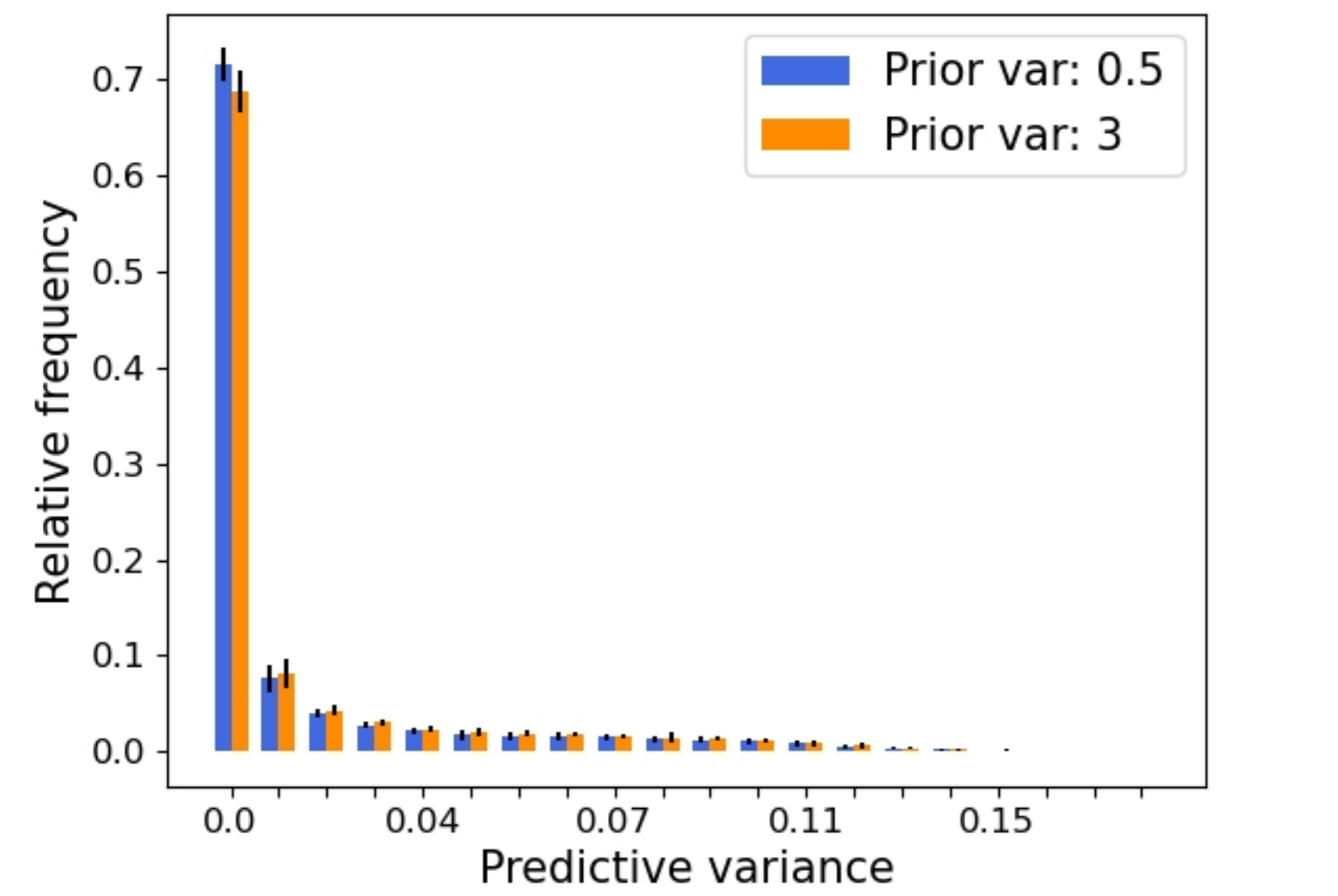}}
    \caption{ 
    Mean variances for models trained with a prior with variance 0.5 (blue) or 3 (orange) on MNIST, error bars indicate 3-fold standard deviations.
    Top: variance of the  weights between the two hidden layers under the mixing distribution. Bottom: Predictive variance of the mixture components.
    }
    \label{fig:prior_variances}
\end{figure}

For our analysis, we trained models with both objectives and $\lambda=1$  with different priors, 
namely matrix variate normal distributions which relate to multivariate normal distributions $\mathcal{N}(0, \sigma^2 I)$ 
with $\sigma^2 \in \{0.5, 1, 1.5, 3 \}$. 

Our results in table~\ref{tab:accuracy_priors} show, that the test accuracy is slightly decreased with increasing prior variance. 
On the other hand, as shown in figure~\ref{fig:prior_variances} (top row) the variance of the  
mixing distribution is increased 
with increasing prior variance, as expected (hypothesis one, (i)).
Corresponding to our second hypothesis, the variance of the mixing distribution of the VI-based model seems to be stronger affected by the choice of the prior variance than the mixture trained with the ML objective.
Interestingly though, only for the  ML objective the increased mixture variance translated to notably higher predictive variances, while the effect is less pronounced for VI, as shown in figure~\ref{fig:prior_variances}, bottom row. 
Additional results investigating the variance of the mixing distribution (figure~\ref{fig:app_prior_variance}) can be found in the appendix.

\subsection{The impact of $\lambda$} 
\label{subsec:lambda_impact}

In this section  we examine the effect of the choice of $\lambda$ for a fixed prior with unit variance.
To do so, we trained models
based on $\mathcal{L}_{ML}$ and $\mathcal{L}_{VI}$ with
$\lambda \in \{  1, 0.1, 0.01, 1\text{e-}3, 1\text{e-}4 \}$.
In table~\ref{tab:accuracy_lambdas}, we show the
resulting mean test accuracies
over 10 trials with different random initialization and their standard deviation. 
The mean accuracy is the highest for $\lambda = 0.1$ for both objectives and data sets.

\begin{table}[h!]
\caption{Accuracy on test set for different values of $\lambda$.}
\centering
\resizebox{\textwidth}{!}{%
\begin{tabular}{l l |l l l l l } 
\toprule
 &   & 1  & 0.1 & 0.01 & 1e-3 & 1e-4 \\ 
\midrule 
MNIST &  VI  &0.974 $\pm 0.78\text{e-}3 \enspace$ & 
\textbf{0.985} $\pm 0.74\text{e-}3 \enspace$ & 
0.982 $\pm 1.78\text{e-}3 \enspace$& 
0.979 $\pm 1.78\text{e-}3 \enspace$& 
0.979 $\pm 1.25\text{e-}3 \enspace$\\
&ML \ & 0.973 $\pm 0.97\text{e-}3$ & 
\textbf{0.983} $\pm 0.84\text{e-}3$ & 
0.982 $\pm 1.23\text{e-}3$ & 
0.979 $\pm 1.63\text{e-}3$ & 
0.979 $\pm 1.50\text{e-}3$   \\
\hline
FMNIST \ &VI &0.854 $\pm 2.28\text{e-}3$ & 
\textbf{0.895} $\pm 1.16\text{e-}3 $& 
0.890 $\pm 2.87\text{e-}3$ & 
0.886 $\pm 2.01\text{e-}3 $& 
0.888 $\pm 2.94\text{e-}3$ \\
&ML &0.860 $\pm 2.53\text{e-}3$ & 
\textbf{0.894} $\pm 1.25\text{e-}3$ & 
0.888 $\pm 3.42\text{e-}3$ & 
0.887 $\pm 2.42\text{e-}3$ & 
0.886 $\pm 2.25\text{e-}3$\\
\bottomrule
\end{tabular}%
}
\label{tab:accuracy_lambdas}
\end{table}

\begin{figure}[h!]
\vspace{-1.3cm}
    \centering
    \subfloat[ML, MNIST ]{\includegraphics[width=0.48\linewidth]{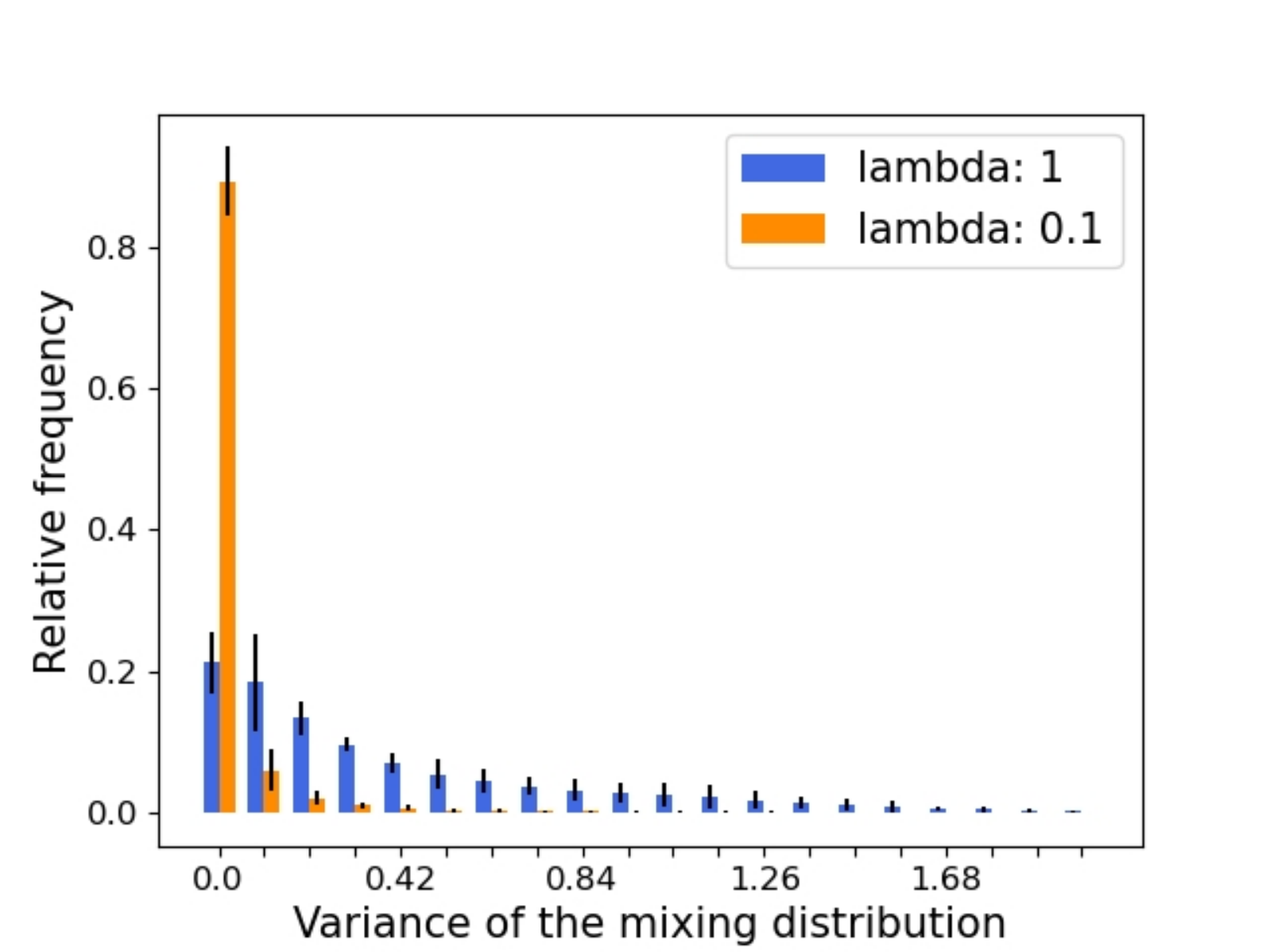}}
    \subfloat[VI, MNIST]{\includegraphics[width=0.48\linewidth]{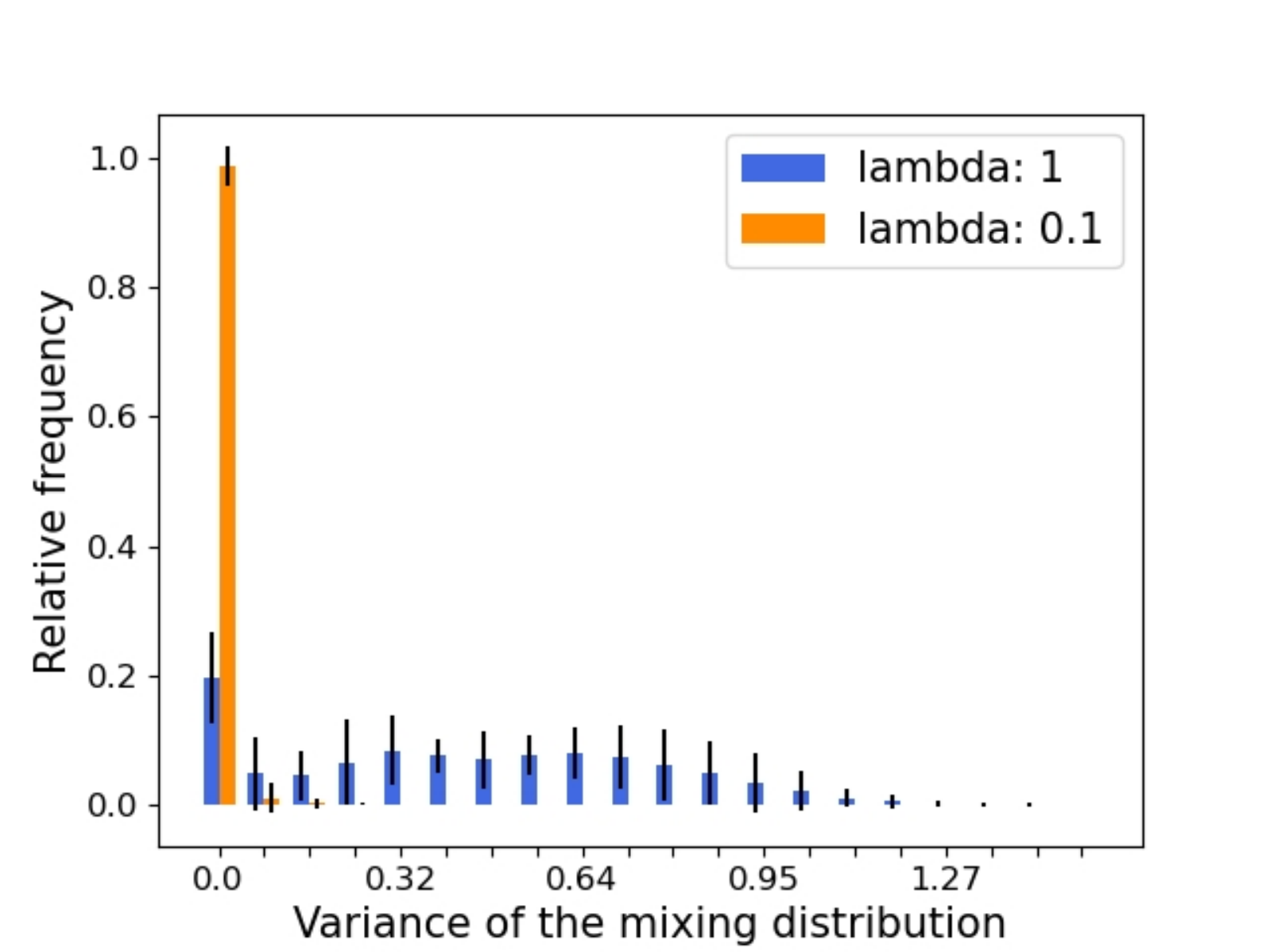}}
    \caption{Variance of the weights between the two hidden layers under the mixing  distribution of  ML- and VI-based models trained on MNIST with $\lambda = 1$ and $\lambda = 0.1$.  We report the mean over 10 trials, error bars indicate 3-fold standard deviations.}
\label{fig:mixing_variances}
\end{figure}

Next, we investigate the effect of $\lambda$ on the variance of the weights between the two hidden layers under the mixing distribution. 
As shown in
figure~\ref{fig:mixing_variances}, $\lambda= 0.1$ led to a significant smaller variance in the 
mixing distribution than $\lambda=1$. This corresponds to our expectations formulated in hypothesis one (ii), where we connected the impact of the prior to the weight of the KLD term.

\begin{figure}[h!]
    \centering
    \subfloat[MNIST $\lambda=1$]{\includegraphics[width=0.48\linewidth]{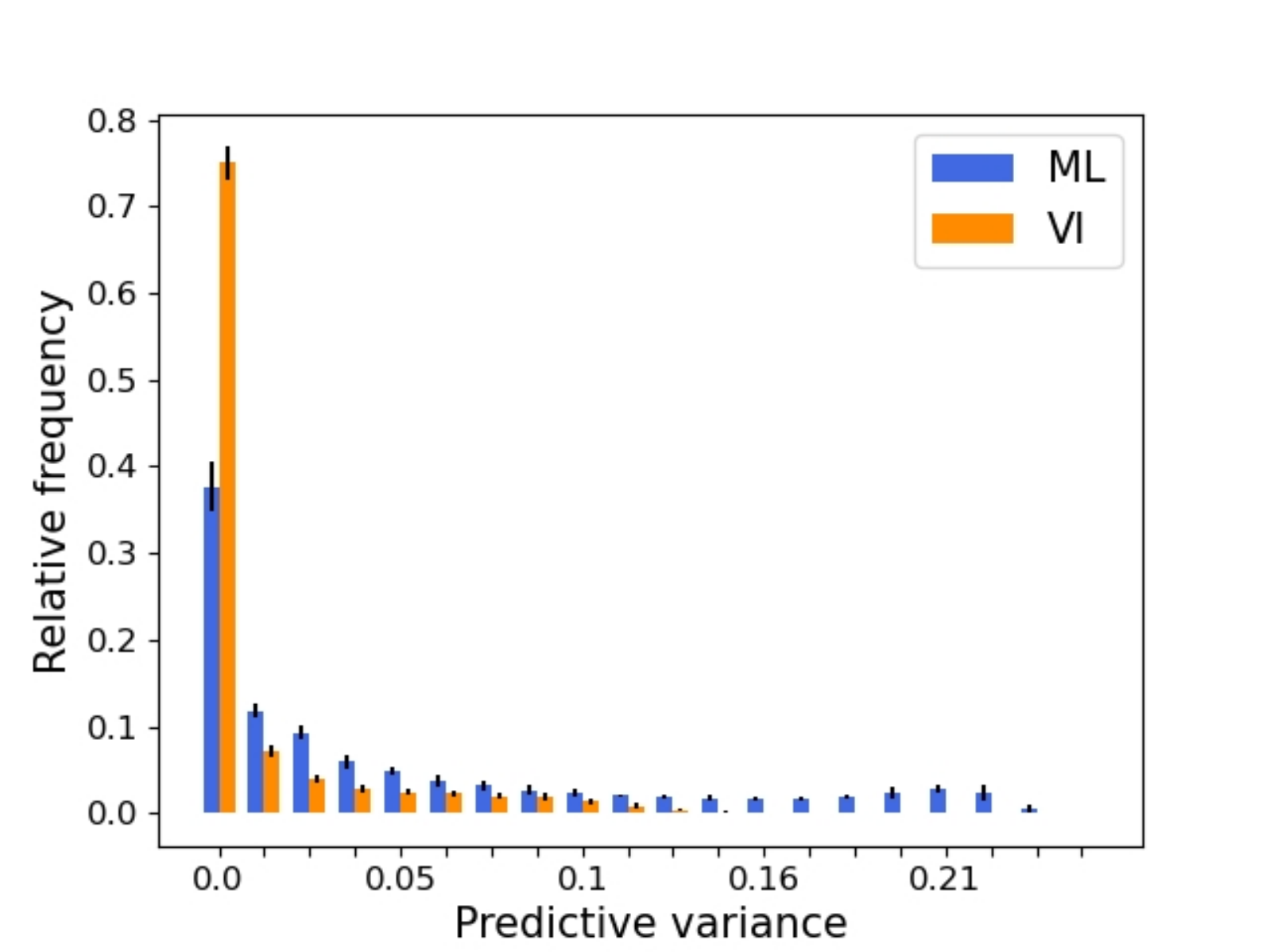}}
    \subfloat[FMNIST $\lambda=1$]{\includegraphics[width=0.48\linewidth]{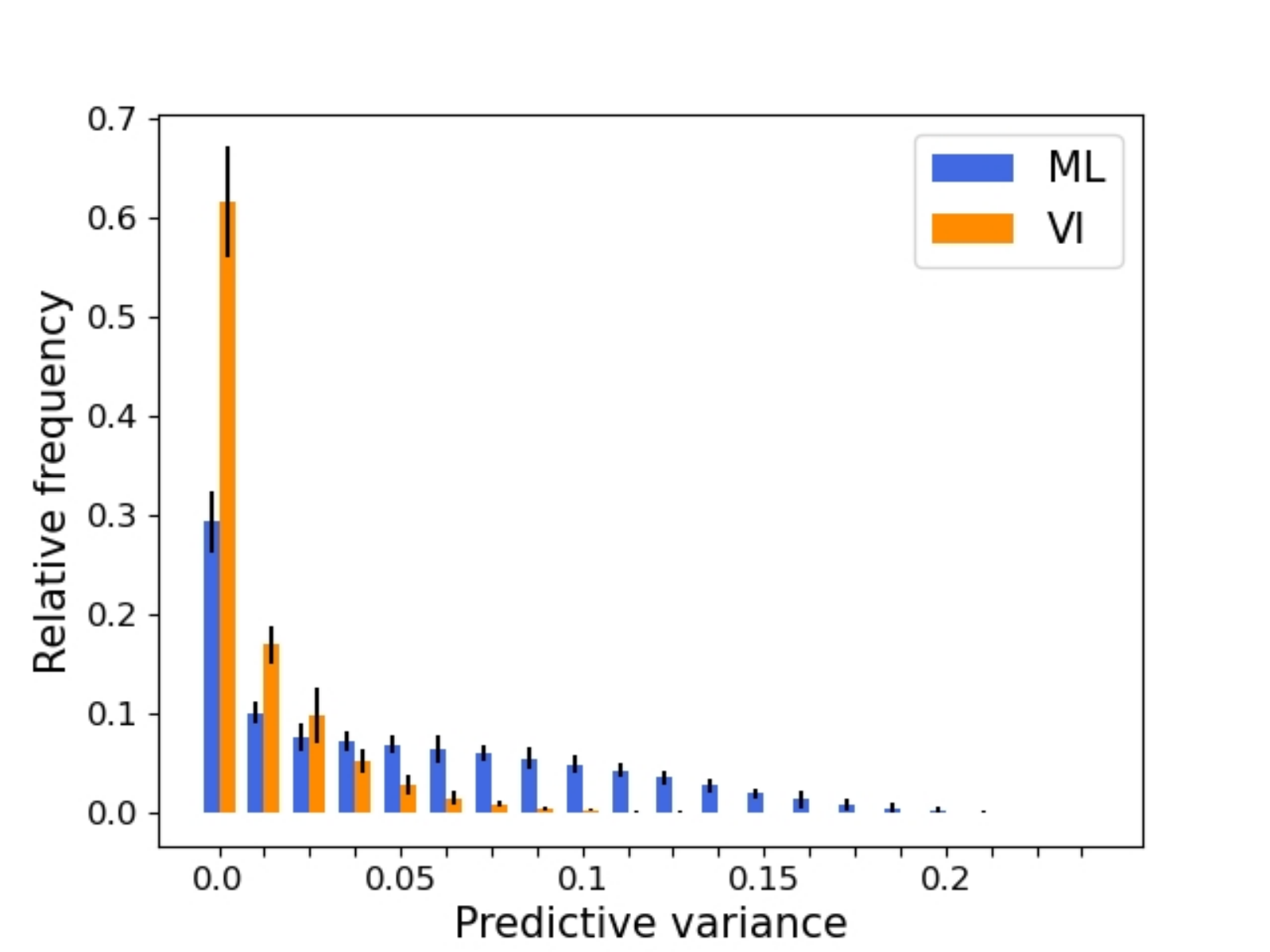}}
    \qquad
    \subfloat[MNIST $\lambda=0.1$]{\includegraphics[width=0.48\linewidth]{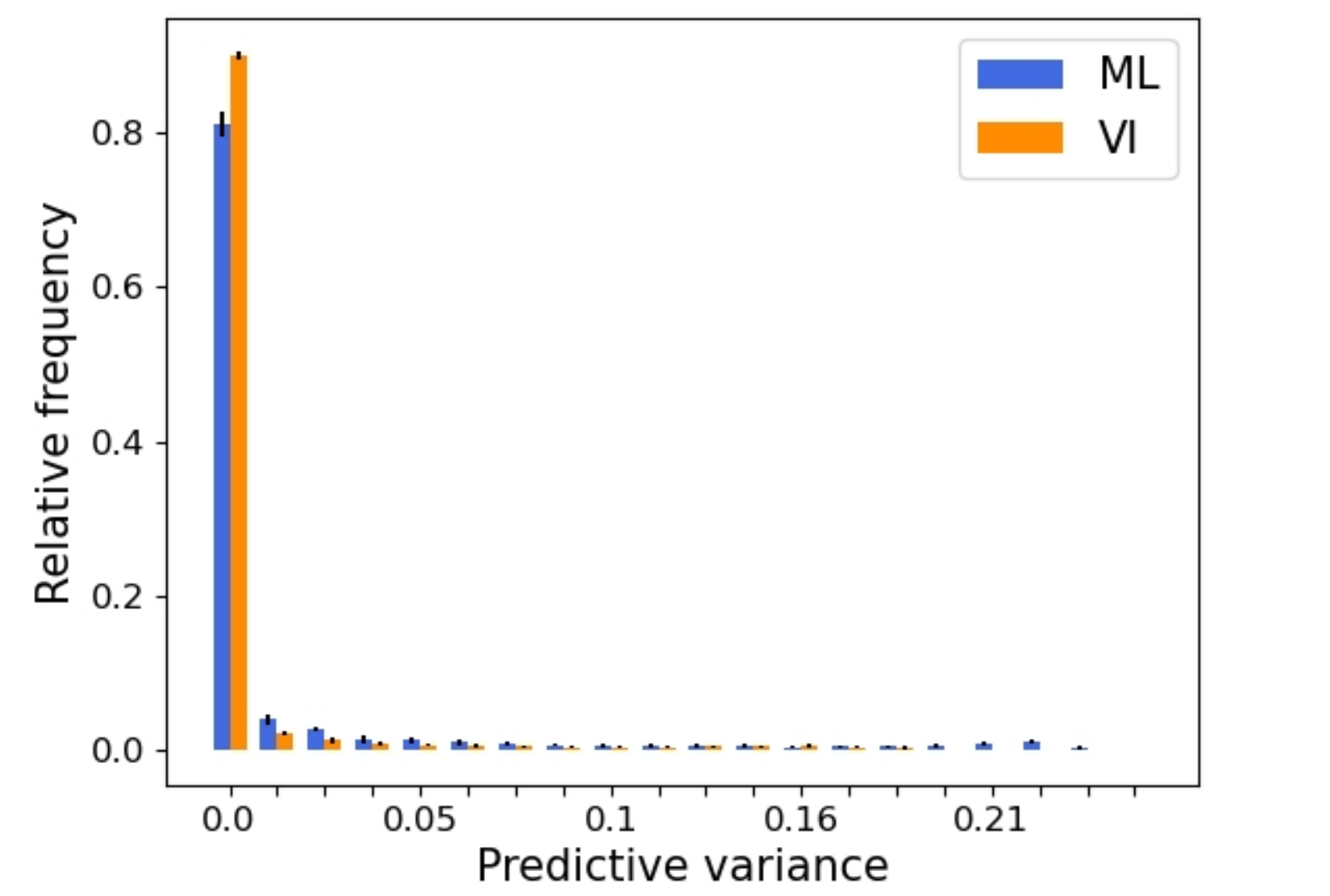}}
    \subfloat[FMNIST $\lambda=0.1$]{\includegraphics[width=0.48\linewidth]{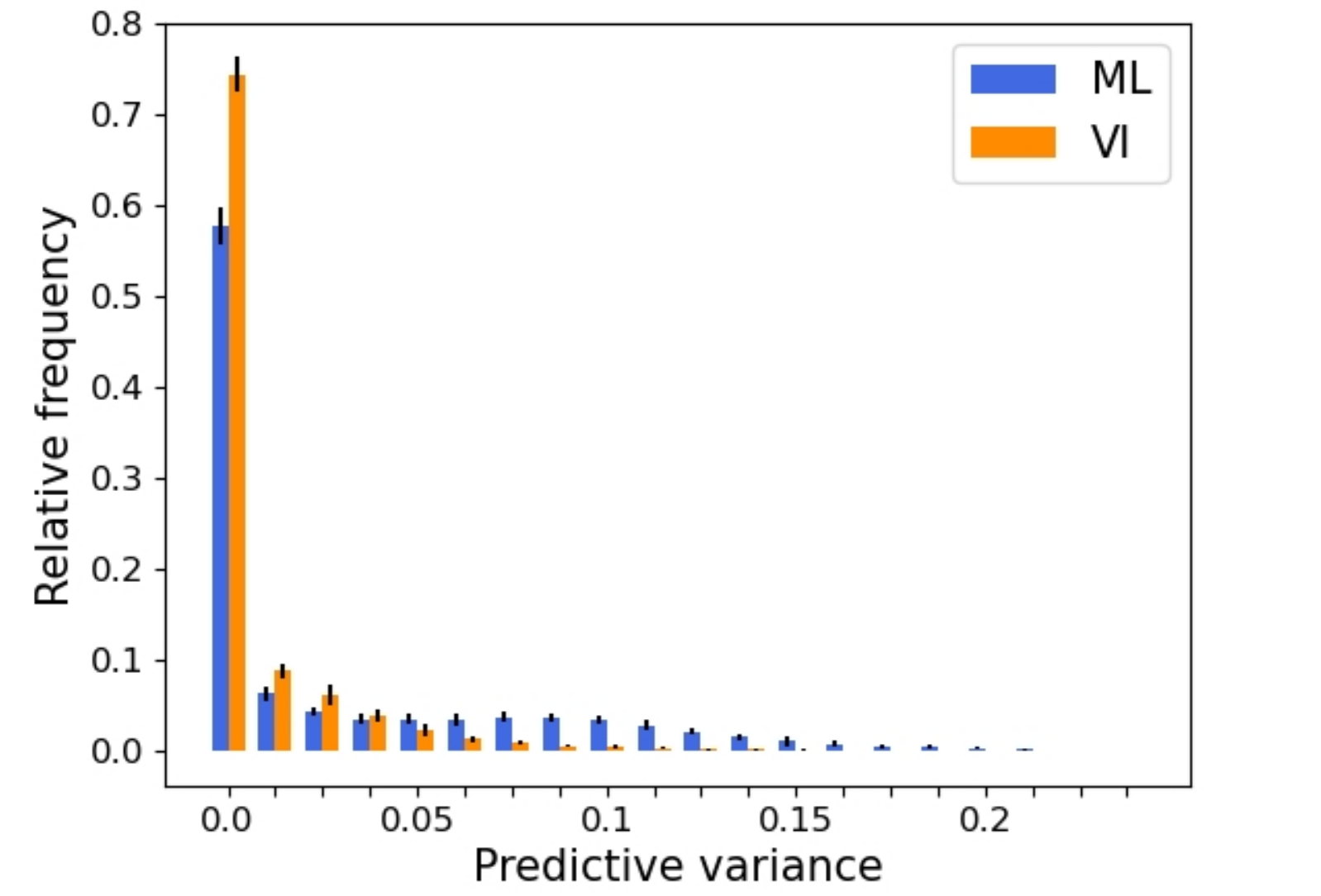}}
    \caption{Predictive variance between the mixture components of ML- and VI based models for MNIST and FMNIST, 
    top: $\lambda = 1$, bottom: $\lambda =  0.1$. We report the mean over 10 trials. The error bars indicate 3-fold 
    standard deviations.
    }
    \label{fig:variances}
\end{figure}

Based on this observations, we investigate if the smaller variance in the mixing distribution also leads to a smaller predictive variance. 
Figure~\ref{fig:variances} compares the predictive variances  on the test set of models trained with $\mathcal{L}_{ML}$ or  $\mathcal{L}_{VI}$ and $\lambda   = 1$ (upper row) with those trained with $\lambda = 0.1$ (lower row). 
These histograms indicate that the smaller weight of the KLD term led indeed to a decrease in the predictive variance of both models. 
Moreover, they show that
the ML objective led to higher predictive variance than the VI objective, which could again be explained by our third hypothesis. 

In comparison to the results of section~\ref{subsec:priors}, it seems that the strength of the 
KLD term has a higher impact on the variance of the final mixing distribution and the predictive variance than the prior variance itself.

\subsection{Detecting miss-classification}
\label{subsec:test}
To investigate if the increase in predictive variance of the model trained with the ML objective 
compared to the model trained with the VI objective can be used to identify miss-classification, 
we examine the predictive variance for 
correctly and wrongly classified test examples separately.
In this 
analysis we focus on models trained with $\lambda=1$ and a prior with variance one, since this led to good results in our previous experiments.
\begin{figure}[h!]\label{fig}
    \centering
    \subfloat[ML, MNIST]{\includegraphics[width=0.48\linewidth]{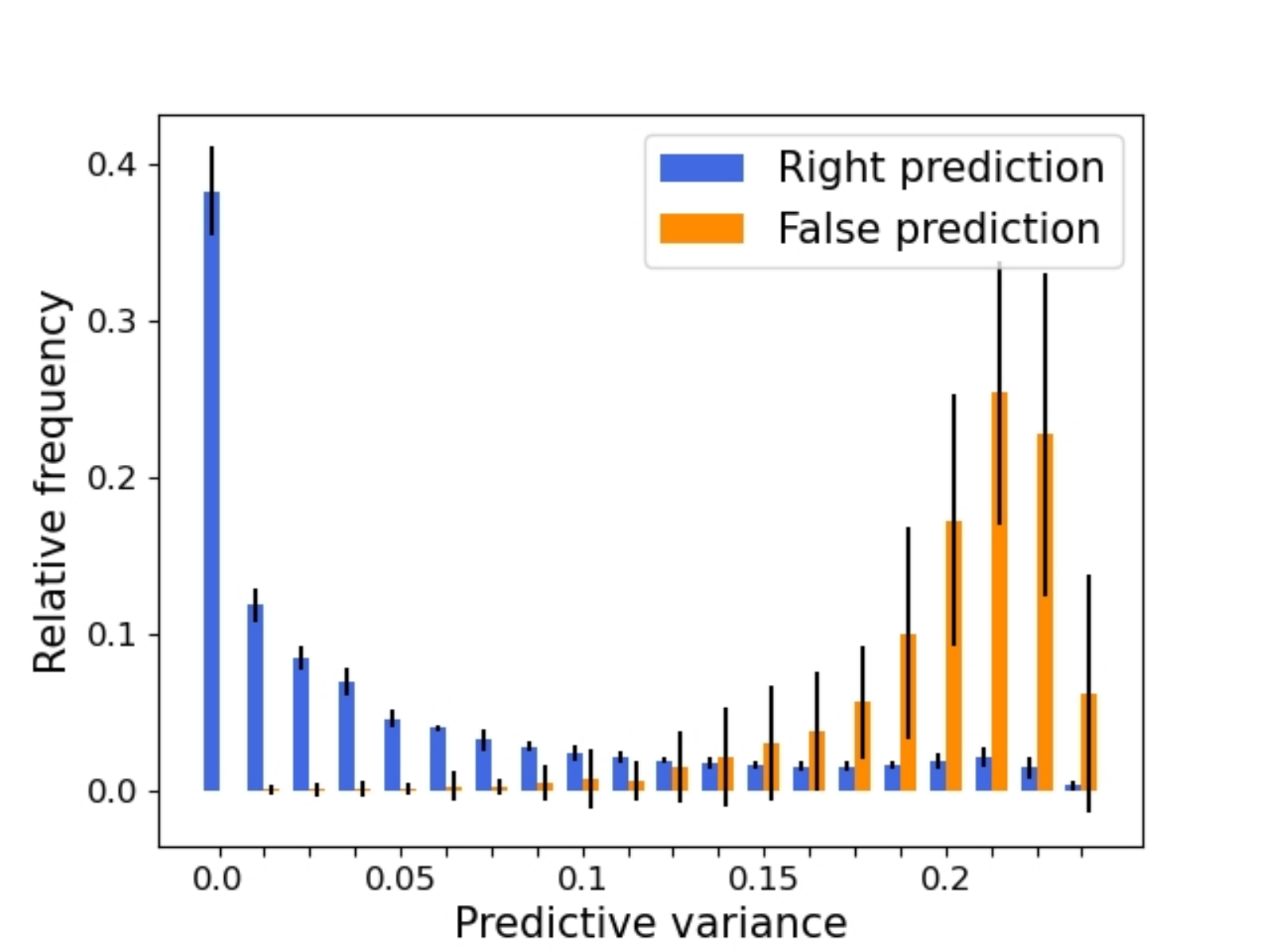}}
    \subfloat[VI, MNIST]{\includegraphics[width=0.48\linewidth]{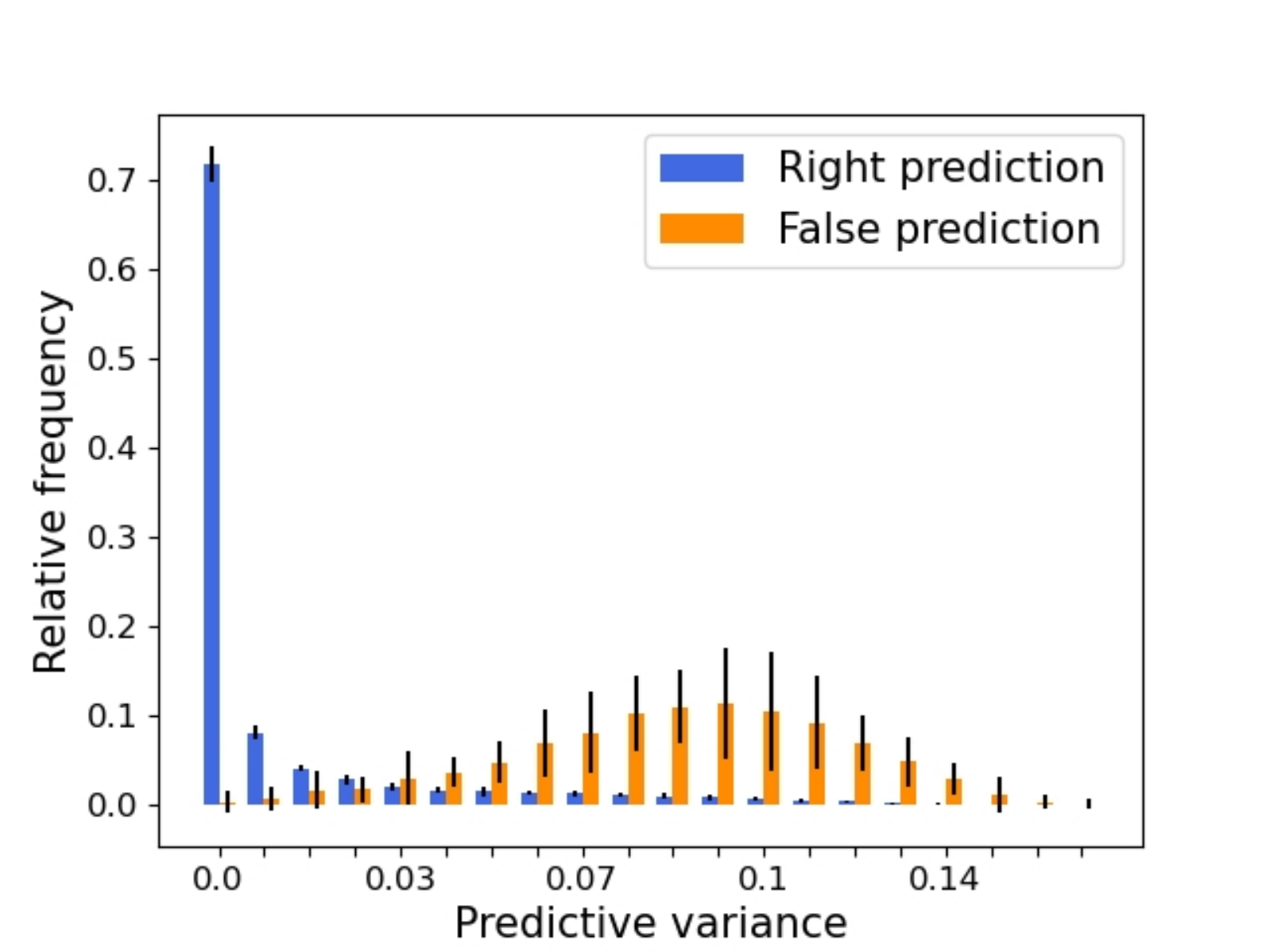}}
    \qquad
    \subfloat[ML, FMNIST  ]{\includegraphics[width=0.48\linewidth]{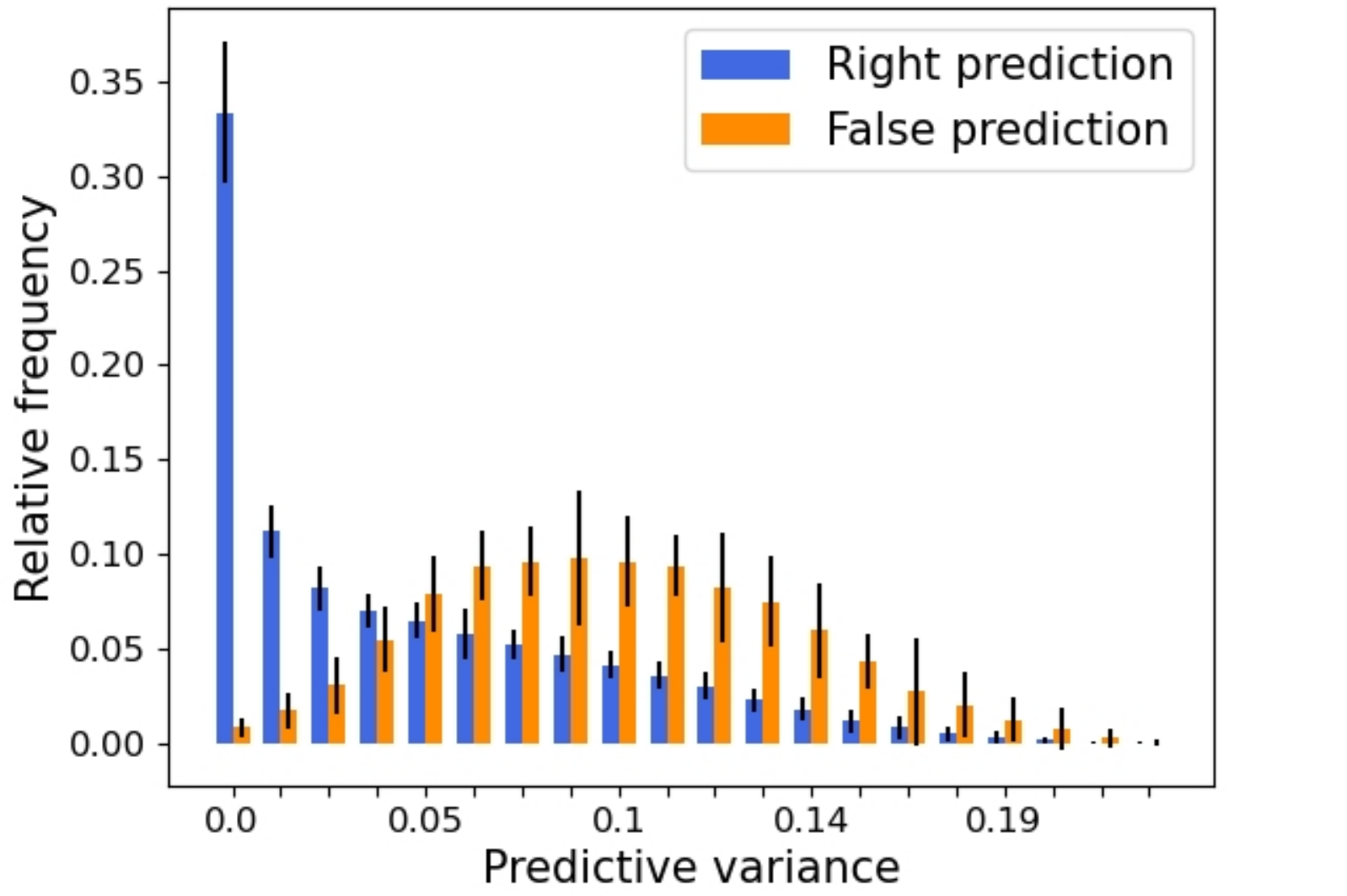}}
    \subfloat[VI, FMNIST]{\includegraphics[width=0.48\linewidth]{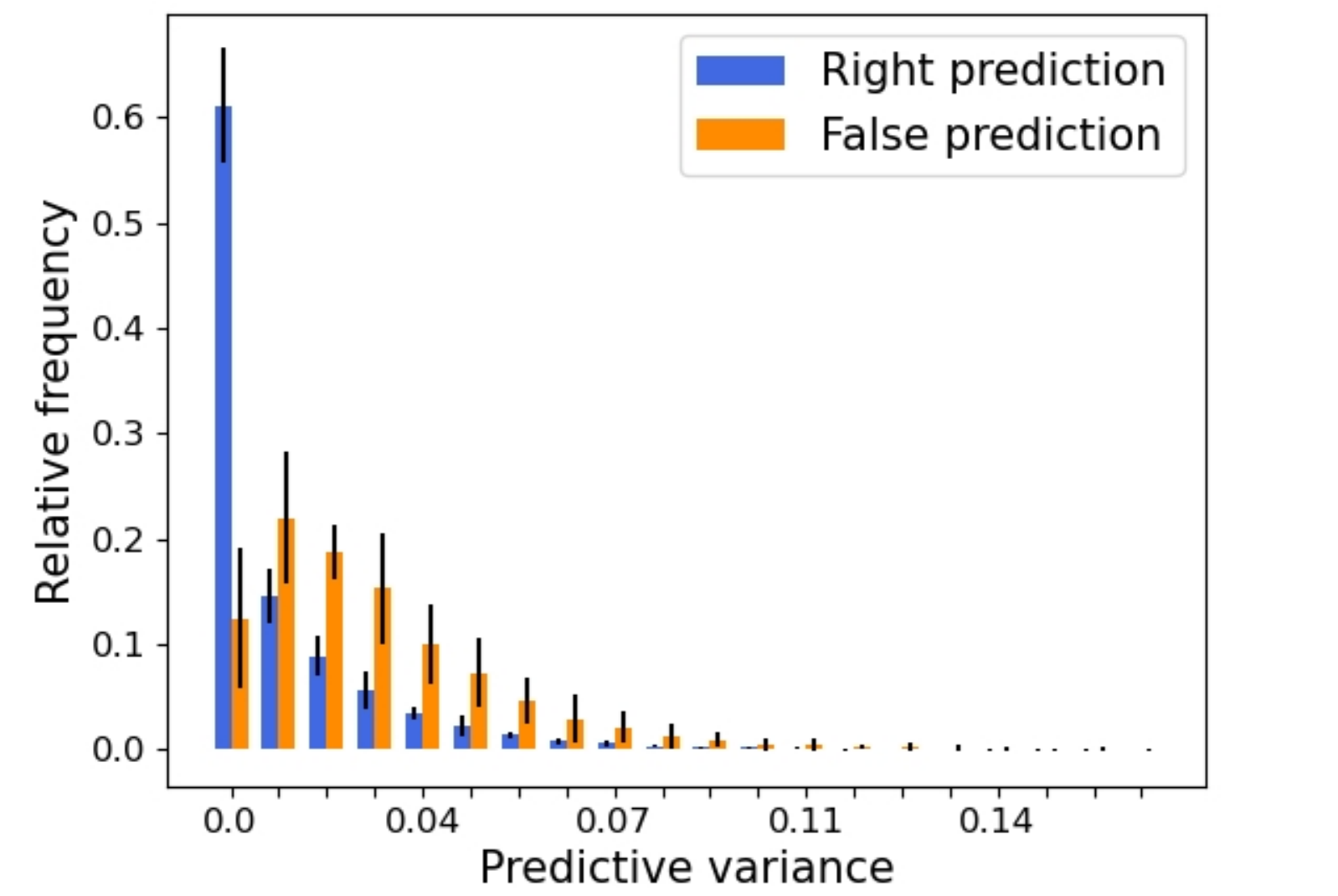}}
    \caption{
    Predictive variance between the mixture components of ML- and VI-based models 
    for correctly and wrongly classified test samples
    on MNIST and FMNIST.
    }
    \label{fig:test_variance}
\end{figure}

Figure~\ref{fig:test_variance} shows that 
both models have a larger predictive variance for wrongly classified examples. 
However, the difference between correctly and wrongly classified examples is more distinct for the ML-based model, which would allow for a better uncertainty based detection of miss-classification. 
Similar results (see  figure~\ref{fig:test_entropy} in the appendix) can be derived when 
investigating
the entropy of the predictive distribution of the mixture model.

\subsection{Detecting out-of-distribution data}
\label{subsec:ood}

In this section we examine 
if the increased predictive variance of the ML-based model can also be exploited for the detection of \textit{out-of-distribution} (OOD) data. For this purpose we used the first 10,000  training samples from the the \textit{"notMNIST"} data set\footnote{Available under http://yaroslavvb.blogspot.com/2011/09/notmnist-dataset.html .} (which consists of 18,724 images of letters from A to J in different font styles and with the same resolution as MNIST and FMNIST) and investigated if 
they can be classified  as OOD data, based on an increased predictive variance compared to the test data. We also included the entropy (as calculated in eq.~\eqref{eq:entropy}) as an additional measure of uncertainty in this analysis. That is, we classify  a sample as true data point if the predictive variance/entropy over the classes under the corresponding model is smaller than a given threshold and as OOD if the predictive variance/entropy is larger.

We compare the performance of the models trained with the $\mathcal{L}_{ML}$ and  $\mathcal{L}_{VI}$ objective 
(with $\lambda=1$ and a prior with unit variance)
in terms of
the \textit{area under the receiver operating characteristic curves} (AUROC) in table~\ref{tab:auroc_ood}.
If the classification is based on the predictive variance 
both models show a similar performance.
However, the ML-based model allows for better entropy based OOD detection than the VI-based model.

\begin{table}[h!]
\caption{AUROC scores for distinguishing between test samples and OOD data.}
\centering
\begin{tabular}{l |l l| l l } 
\toprule
    &MNIST && FMNIST & \\
    & ML  & VI & ML  & VI\\ 
\midrule 
Predictive variance $\enspace $ & 0.953 $\pm 0.41\text{e-2}  \enspace $
& 0.955 $\pm 0.70\text{e-2} \enspace $
& 0.873 $\pm 0.78\text{e-2} \enspace $
& \textbf{0.876}$\pm 1.04\text{e-2} \enspace $\\
Entropy  & \textbf{0.976} $\pm 0.41\text{e-2}  $
& 0.932 $\pm 0.87\text{e-2} $& 
0.814 $\pm 0.82\text{e-2}$ & 
0.756 $\pm 1.94\text{e-2}$\\
\bottomrule
\end{tabular}
\label{tab:auroc_ood}
\end{table}

\subsection{Robustness against adversarial attacks}
\label{subsec:stronger_attacks}
Next, we examine the effect of the different objectives on the
robustness of the models when attacked by adversarial examples calculated based on the projected gradient method \cite{Madry} implemented in cleverhans \cite{papernot2018cleverhans} with increasing perturbation strength.

\begin{figure}[t!]\label{fig}
    \centering
    \subfloat[ML, MNIST]{\includegraphics[width=0.48\linewidth]{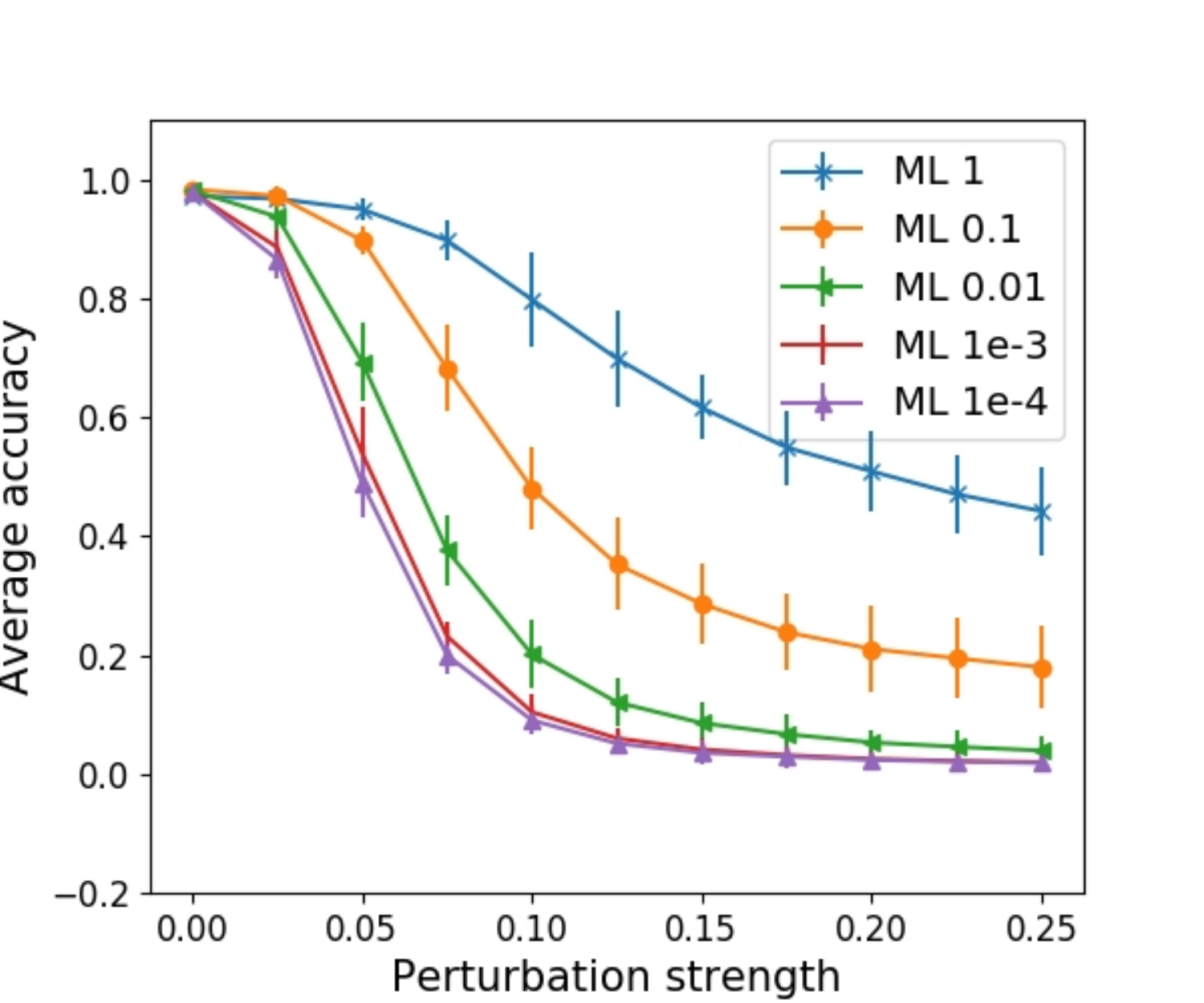}}
    \subfloat[ VI, MNIST]{\includegraphics[width=0.48\linewidth]{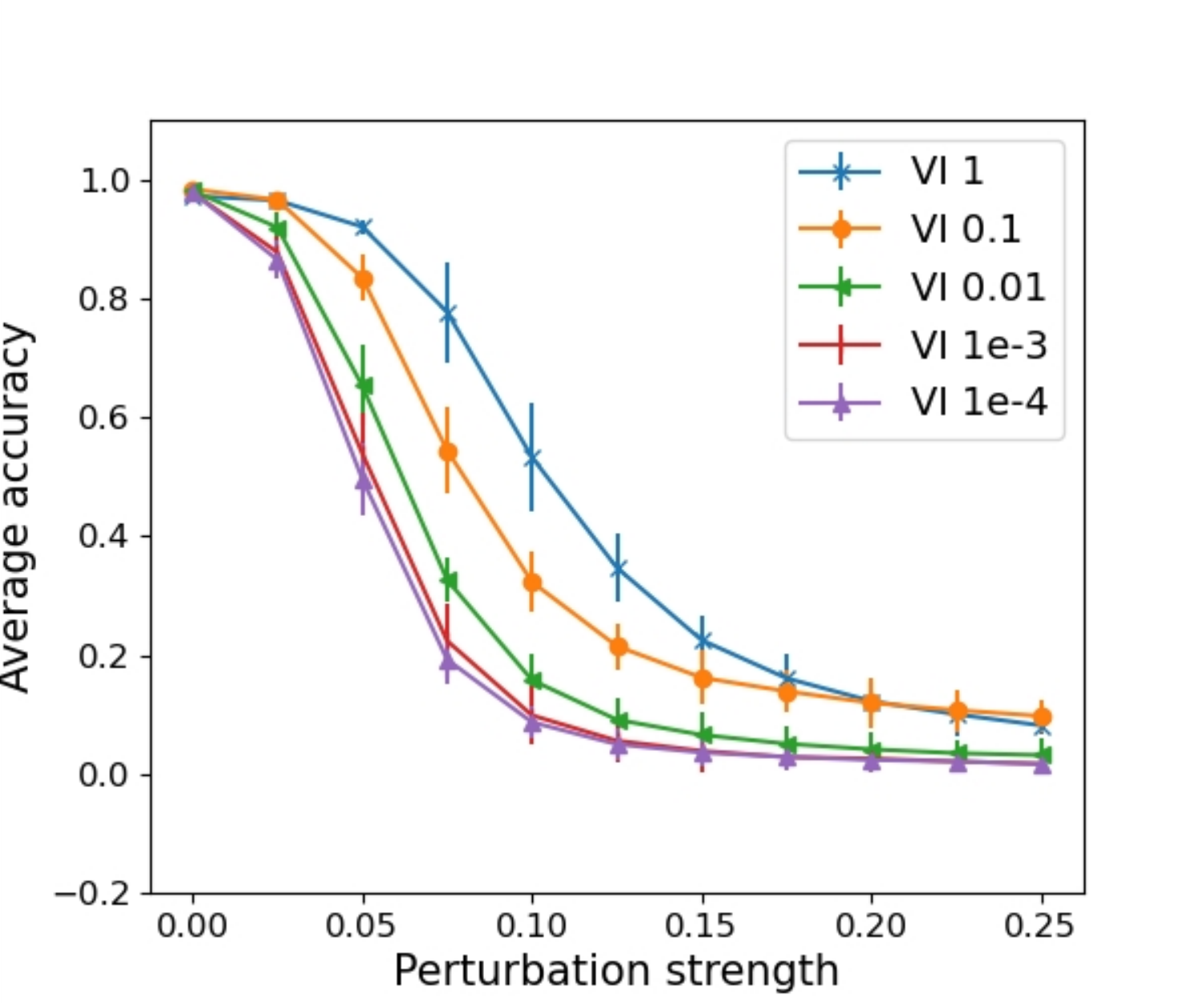}}
    \qquad
    \subfloat[ ML, FMNIST ]{\includegraphics[width=0.48\linewidth]{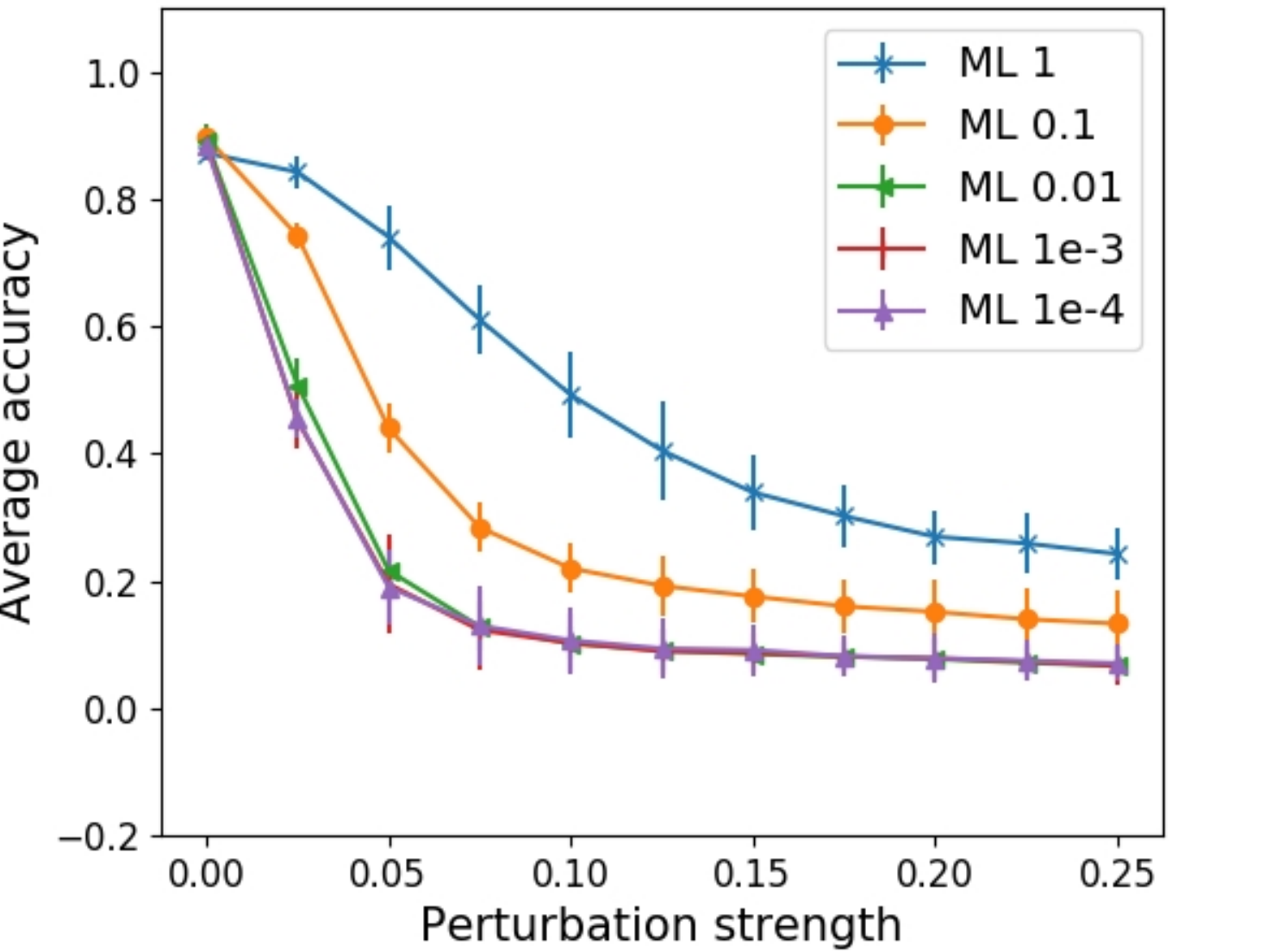}}
    \subfloat[ VI, FMNIST ]{\includegraphics[width=0.48\linewidth]{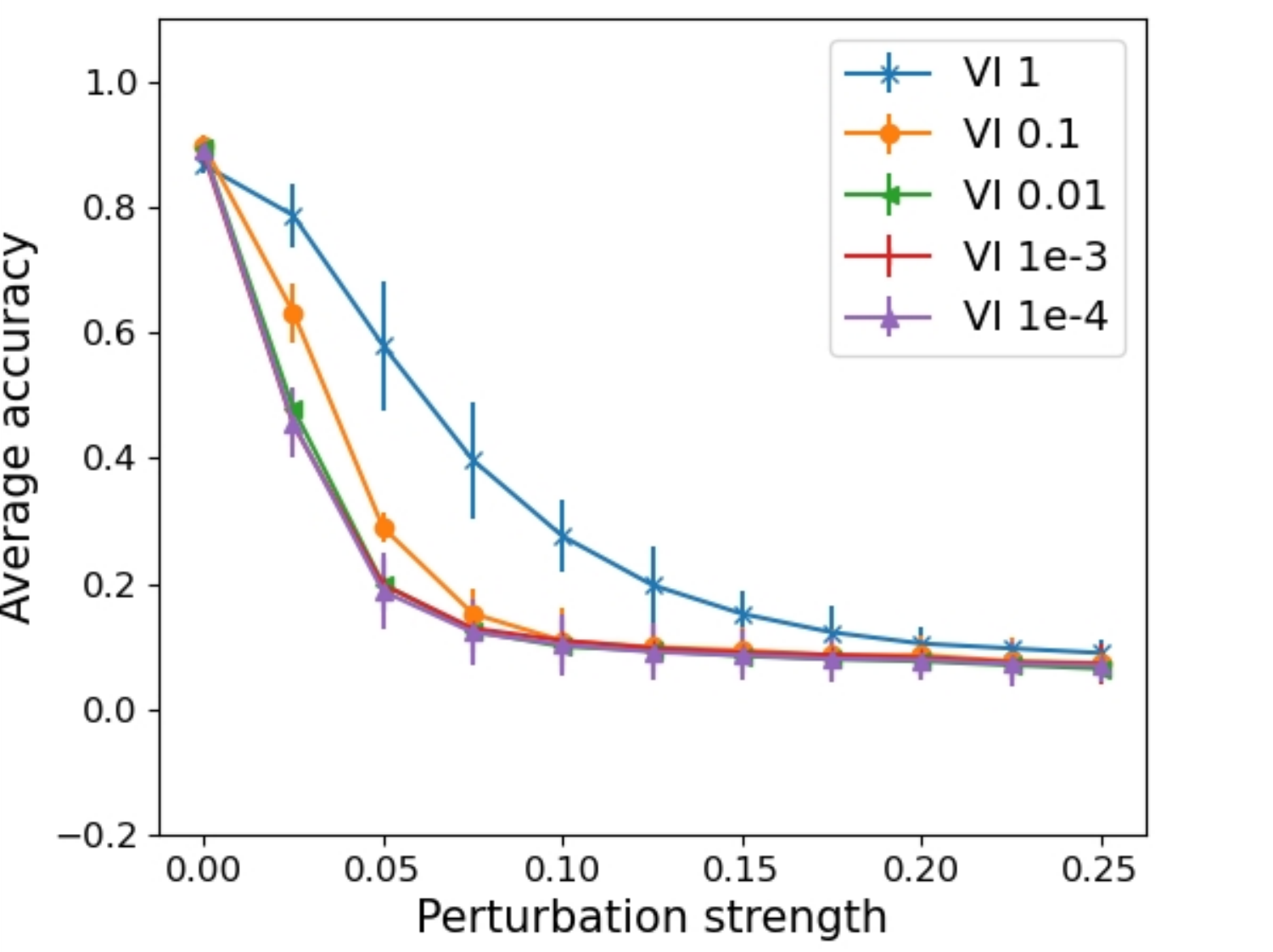}}
    \caption{  Average classification accuracy over 1,000 adversarial examples ($3$-fold standard deviation indicated by error bars) in dependence of the perturbation strength,
    for ML- and VI- based models with varying values for $\lambda$ (compare eq.~\eqref{eq:VI-objective} and \eqref{eq:ML-objective}),
    top: MNIST, bottom: FMNIST.
    }
    \label{fig:lambda_accuracy}
\end{figure}

First, we investigate the impact of different values of $\lambda$ and use one gradient sample to calculate the adversarial examples. 
The results in figure~\ref{fig:lambda_accuracy} show that lower values of $\lambda$ lead to models less robust to adversarial examples. Largest robustness was observed for $\lambda=1$ which however led to models less accurate on the test set compared to models trained with $\lambda=0.1$ (compare table~\ref{tab:accuracy_lambdas}).
We refer to this as the  accuracy-robustness trade-off\footnote{We note that the deviation of some kind of measure that would help hyperparamter optimization w.r.t.~the accuracy-robustness trade-off would be valuable for the community and might be an interesting topic for future work.
}.

While smaller values of $\lambda$ reduce the robustness w.r.t.~adversarial examples of both models, larger values of $\lambda$ seem explicitly advantageous for the robustness of ML trained mixtures. 
Both observations 
could be a consequence of the fact, that the predictive variance of stochastic nets relates to the variance of the stochastic gradients adversarial attacks rely on. A higher gradient variance makes adversarial attacks relying on a single gradient sample less efficient which indirectly improves the robustness of the corresponding model \cite{carlini2018obfuscated,Handtuch}. This explains why
the previously observed effect of increasing the variance of the mixing distribution as well as the predictive variance by choosing a larger value of $\lambda$ might translate into an increase in
robustness
\footnote{Interestingly, varying the 
prior variance 
directly had
hardly any effect on the robustness as can be seen in figure \ref{fig:app_prior_accuracy} in the appendix. This fits our previous observation that the choice of $\lambda$ has a larger impact on the predictive variance than the choice of the variance of the prior.}.
Likewise, the advantage of ML over VI trained models 
can potentially be traced back to the larger predictive variance of the ML based models due to the properties of the expectation term (as outlined in our 
third hypothesis). 

\begin{figure}[t!]\label{fig}
\vspace{-0.5cm}
    \centering
    \subfloat[ML, MNIST]{\includegraphics[width=0.48\linewidth]{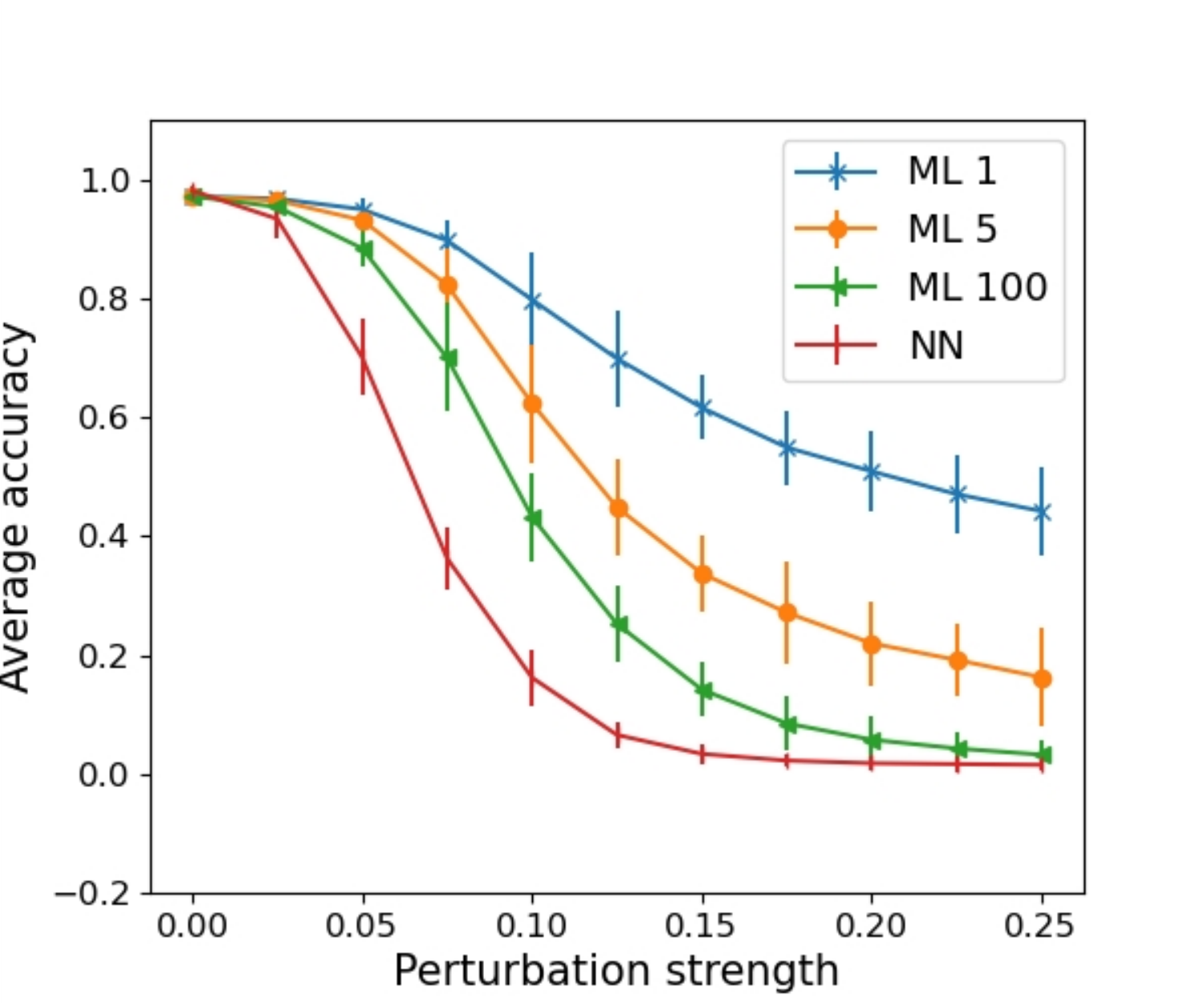}}
    \subfloat[VI, MNIST]{\includegraphics[width=0.48\linewidth]{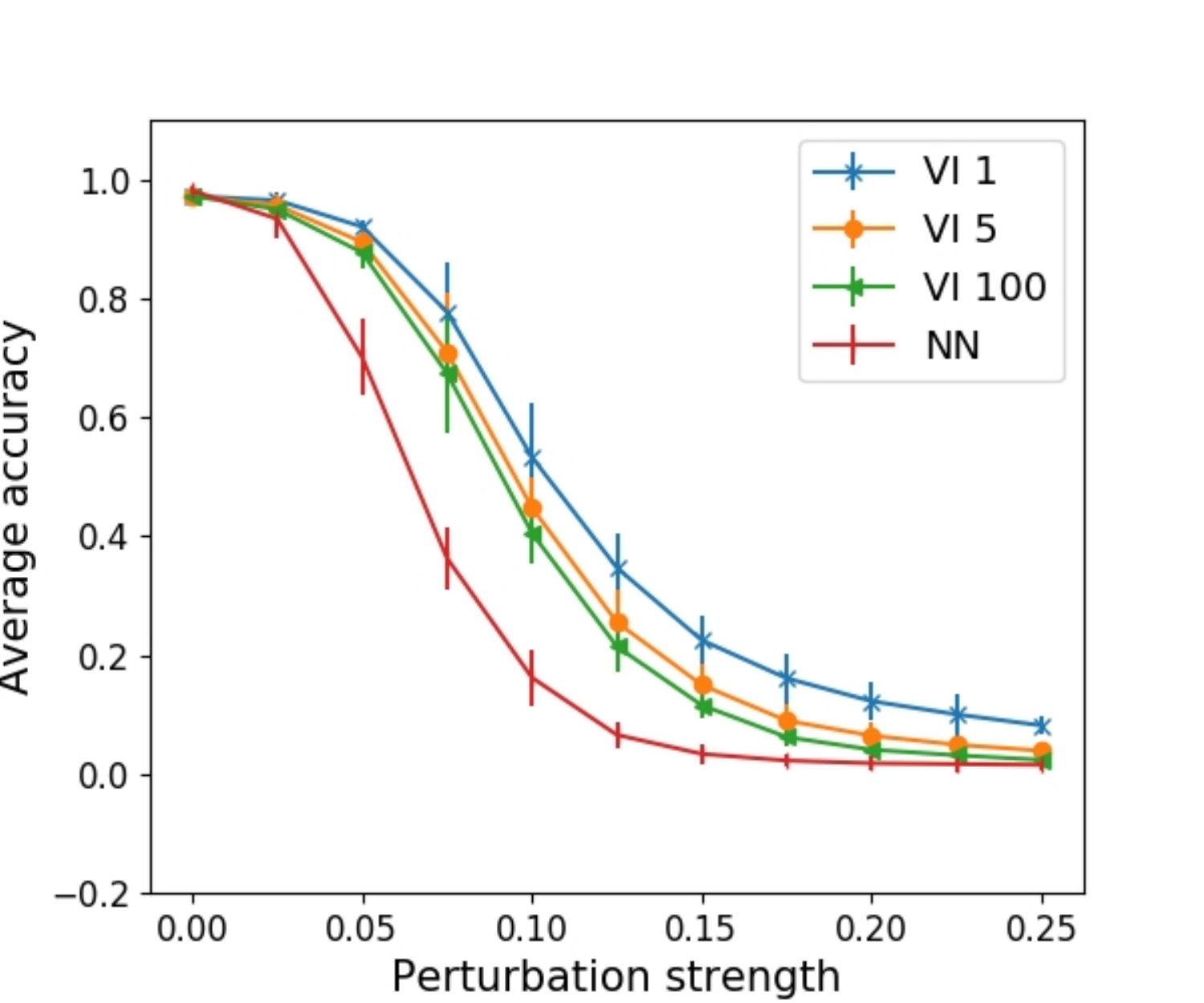}}
    \qquad
    \subfloat[ML, FMNIST]{\includegraphics[width=0.48\linewidth]{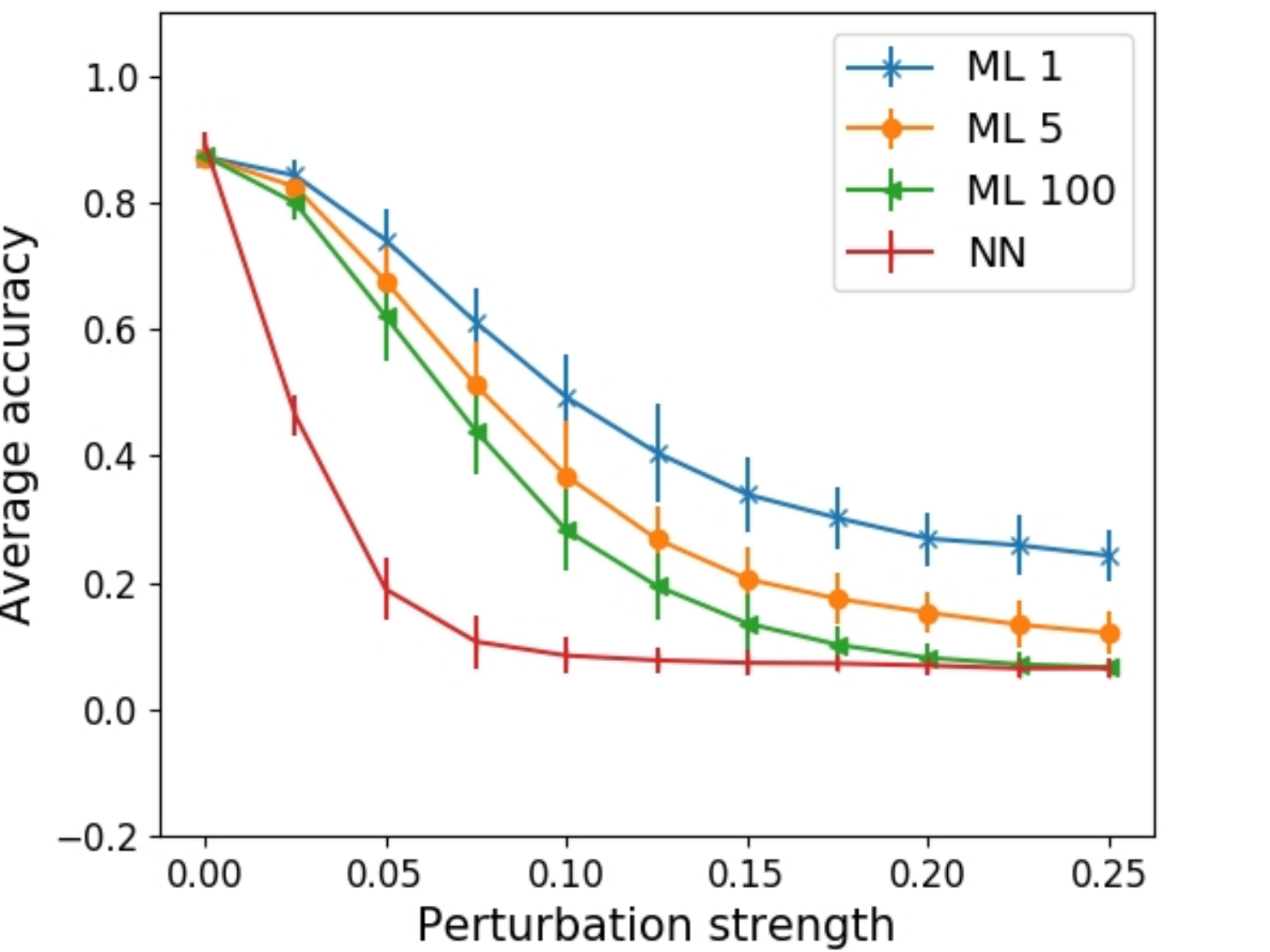}}
    \subfloat[VI, FMNIST]{\includegraphics[width=0.48\linewidth]{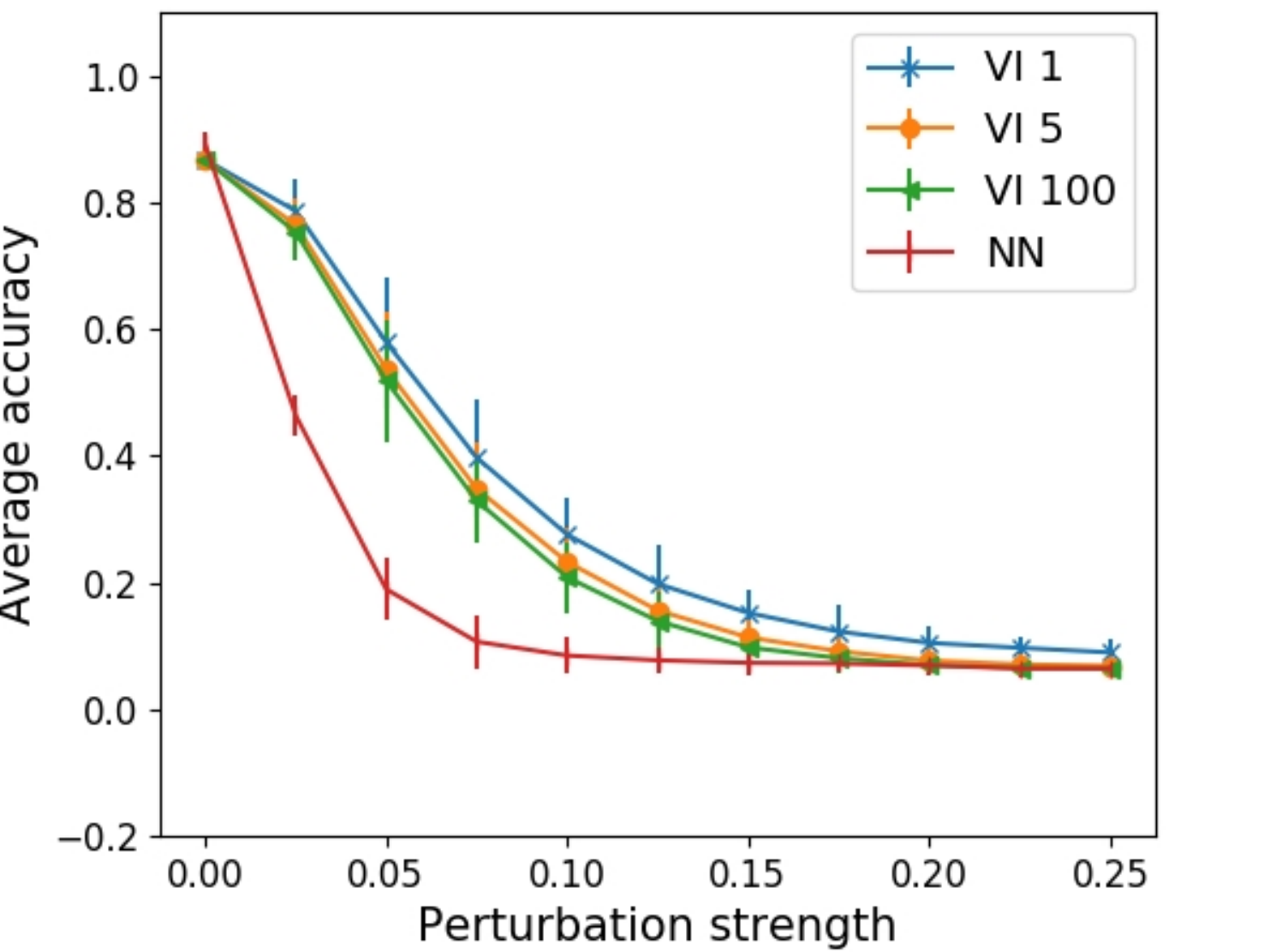}}
    \caption{Average accuracy for stronger adversarial examples 
    based on an increased sample size (5 or 100) for estimating
    the stochastic gradients. The accuracies for the one-sample attack and a standard feedforward network (NN) are also shown for comparison.}
    \label{fig:stronger_attacks}
\end{figure}

To investigate the impact of the stochasticity of the gradient on the results, we increased the strength of the adversarial attacks on the models trained with $\mathcal{L}_{ML}$ and $\mathcal{L}_{VI}$  (with $\lambda=1$ and prior with unit variance) by approximating the integral over $\theta$ in eq.~\eqref{eq:predictive_distribution} during the calculation of the adversarial example with $S= 5$ or $S= 100$ samples instead of only one. 
The results are shown in figure~\ref{fig:stronger_attacks}.
For a comparison to a deterministic network, we added the accuracy of a feed forward neural network (NN)
with the same layer structure as the component models of the mixtures,  but trained with weight decay with a regularization strength of $1/60000$, 
as a baseline.

The results  show, that for both models
the robustness  to adversarial attacks is indeed decreased when using multiple samples for estimating the gradients during the calculation of the adversarial attack.
However, an increased robustness in comparison to the deterministic network (more specifically, an increased
average accuracy for small perturbations till a  strength of about 0.2) still exists. 

Further, the advantage of the ML objective over the VI objective is relativized when more 
samples are drawn for the adversarial attack. This indicates, that, as speculated before, the higher variance in the gradients induced by the higher predictive variance might be the main reason for the
increased robustness of the ML based model in the previous experiment.

\subsection{Detecting adversarial examples}
\label{subsec:detecting_adv_ex}
In this section we investigate 
the differences in uncertainty based
identification of adversarial examples. For that we focused on the models trained with $\lambda=1$ (and a prior with unit variance)
and the adversarial examples obtained from the projected gradient method with a perturbation strength of $0.25$ and one gradient sample. 
We calculated the averaged AUROC scores over ten trials for the task of
distinguishing between samples from the test set and adversarial examples based on
both uncertainty measures described in section~\ref{sec:uncertainty}, namely entropy and predictive variance. Table~\ref{tab:auroc_adv} shows that for both data sets and with both measures the ML-based model produces higher AUROC scores than the VI-based one. Furthermore,
the AUROC scores based on the entropy 
are higher than those based on the predictive variance. On FMNIST the ML-based model only performs slightly better than random guessing, while the VI-based model even performs worse. 

Only considering the correctly classified test data points as the true class and the wrongly classified adversarial examples (e.g. successful attacks) as the negative class, leads to similar results as shown in the appendix.

\begin{table}[h!]
\caption{Mean AUROC scores and single standard deviation for distinguishing between test samples and adversarial attacks.}
\centering
\begin{tabular}{l |l l| l l } 
\toprule
    &MNIST && FMNIST & \\
    & ML  & VI & ML  & VI\\ 
\midrule 
Predictive variance $\enspace$  &0.734 $\pm 1.90\text{e-2} \enspace$
 & 0.624 $\pm 2.92\text{e-2} \enspace$
  & 0.509 $\pm 3.43\text{e-2} \enspace $& 
  $0.344 \pm 3.69\text{e-2} \enspace$\\
Entropy \ & \textbf{0.742}$\pm 1.83\text{e-2}$
 & $ 0.697 \pm 3.64\text{e-2} $
 &  \textbf{0.537}$\pm 3.18\text{e-2}$&
  0.348 $\pm 3.84\text{e-2}$ \\
\bottomrule
\end{tabular}
\label{tab:auroc_adv}
\end{table}

\section{Discussion and Conclusion}
\label{sec:discussion}
In this work we investigate properties of infinite mixture distributions trained with a maximum likelihood (ML) based objective (which we  augment with a Kullback-Leibler divergence (KLD) 
encouraging the distribution over parameters to stay close to a prior)
in comparison to the properties of BNNs 
trained by variational inference (VI), which form infinite mixtures themselves. 
We perform a first empirical analysis with one network architecture trained with both objectives on the MNIST and FMNIST data sets. The results
show that compared to VI
ML training 
leads to an increased predictive variance. We relate this increase to an inherent property of the objective, which is independent of the choice of the prior and the weight of the KLD term.

The higher predictive variance can be exploited to improve uncertainty based identification of  wrongly classified examples. It does not help directly to identify 
out-of-distribution data, where the ML-based model however displays a higher uncertainty in terms of entropy than the VI-based model.

The robustness against (and uncertainty based identification of) 
adversarial attacks computed based on a single gradient sample is significantly increased for ML-based models
due to
the higher predictive variance. 
This advantage of the ML-based  compared to the VI-based mixture 
reduces but is still observable
if the effect of stochastic gradients is diminished
through multiple sampling during the calculation of the adversarial examples.

In summary, our results indicate that infinite mixtures trained with an ML objective might be advantageous for reliable uncertainty quantification during test time and when adversarial attacked. 
Of course, our  experimental analysis is limited, only allowing for preliminary conclusions. Our findings  need to be confirmed by extended experiments including  different model architectures as well as more complicated data sets.

 Another interesting 
 question for future work would be if the theoretical results from Fushiki \cite{fushiki2005} (where an ML based bootstrap prediction is found to be asymptotically more efficient than a Bayesian prediction under misspecified models) can be 
 transferred to our new objective.

%
%

\bibliographystyle{splncs04}
\bibliography{main_FINAL}
\newpage

\appendix

\section*{Appendix}
\label{sec:Supplement}

\section{
Comparison to finite mixture models:
MC dropout and deep ensembles} \label{sec:infvsfin}
Our work was inspired by the interpretation of BNNs, MC dropout, and deep ensembles as mixture models and the different ways to train such mixtures. However, the 
investigations in the main part of the paper are all limited to infinite mixtures. 
For completeness we  therefore add the results of the finite mixtures based on MC dropout and a deep ensemble to our evaluation. We used a dropout probability of $0.5$ for the dropout model and five 
networks to compose the deep ensemble.
Further, we used the same general network structure as in the infinite mixture case, namely 2 hidden layer each with 128 nodes where the parameters are optimized with ADAM and additionally weight decay with a regularization strength of $1/60000$. 

\begin{figure}[h!]
    \centering
    \vspace{-1.4cm}
    \subfloat[MNIST]{\includegraphics[width=0.48\linewidth]{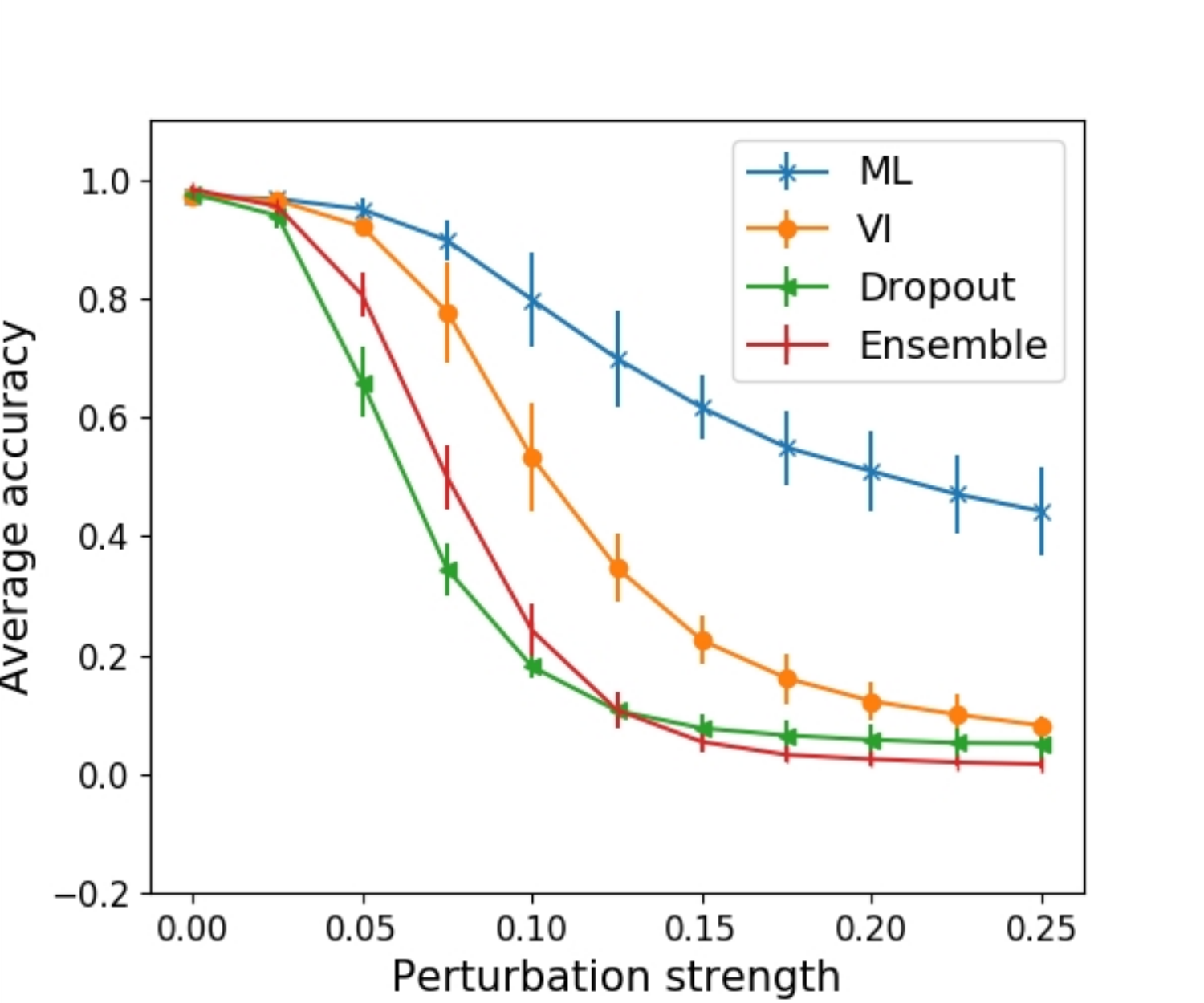}}
    \subfloat[FMNIST]{\includegraphics[width=0.48\linewidth]{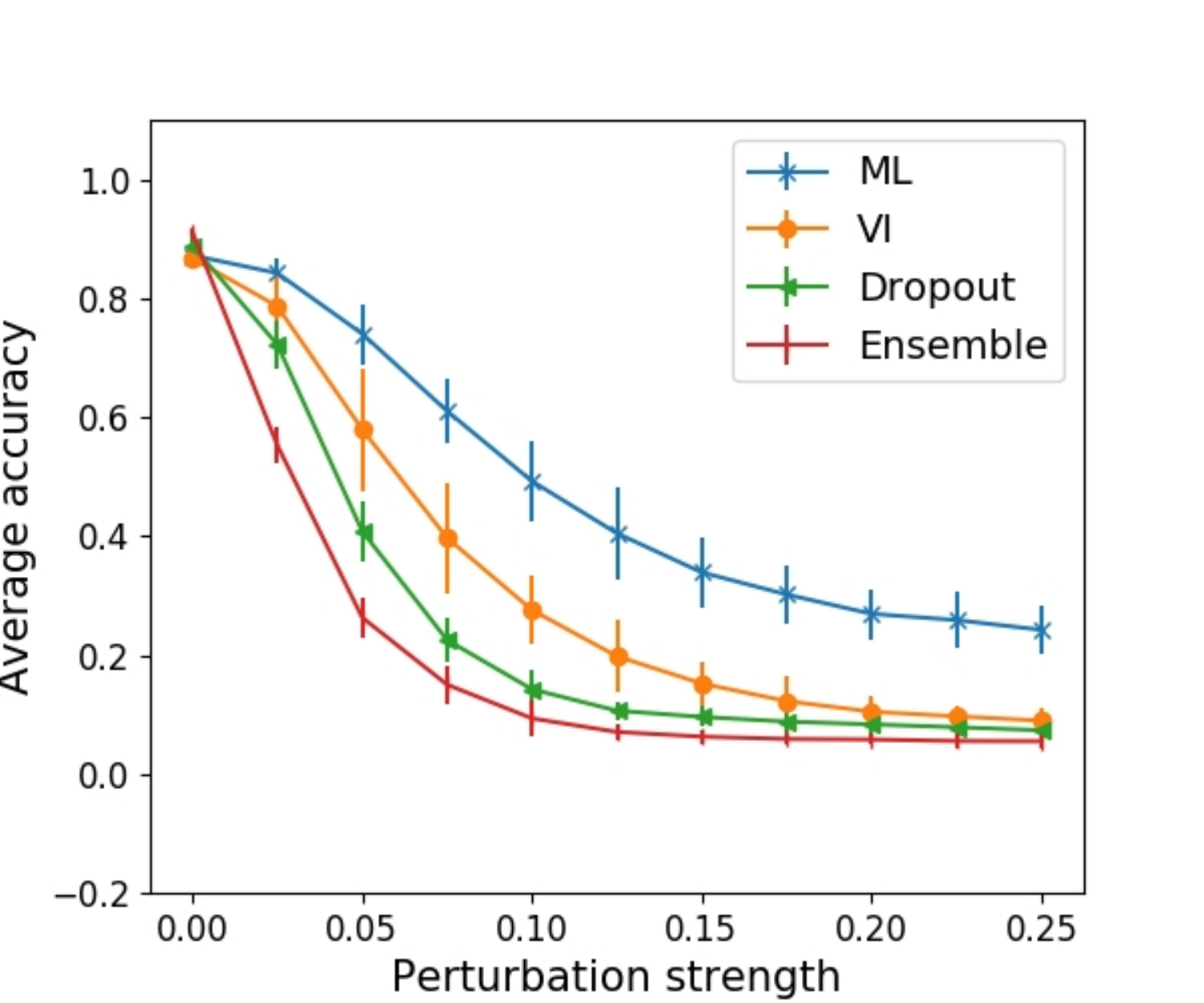}}
    \caption{ 
    Mean classification accuracy
    over 1,000 adversarial examples
    ($3$-fold standard deviation indicated by error bars) in dependence of the perturbation strength
    for ML- and VI based infinite mixtures models trained with $\lambda =1$ 
    in comparison to
    MC dropout and and a deep ensemble.
    }
    \label{fig:diff_models_accuracy}
\end{figure}

While the accuracy 
of both finite mixture models is increased compared to the infinite mixtures (compare table \ref{tab:accuracy_dropoutDE}), their robustness against adversarial examples is not competitive as can be seen in figure \ref{fig:diff_models_accuracy}.

\begin{table}[h!]
\caption{Accuracy on test set for ML and VI optimized infinite mixtures, MC dropout, and 
a deep ensemble.}
\centering
\begin{tabular}{l  |l l l l } 
\toprule
 & ML &VI&  MC dropout  & Deep ensemble \\ 
\midrule 
MNIST & 0.973 $\pm 0.97 \text{e-}3 \enspace$ 
& 0.974 $\pm 0.78 \text{e-}3 \enspace$ 
& 0.978 $\pm 0.63 \text{e-}3 \enspace$ 
&   \textbf{0.985} $\pm 0.63 \text{e-}3$  \\ 
FMNIST & 0.860 $\pm 2.53 \text{e-}3$ 
& 0.854 $\pm 2.28 \text{e-}3$ 
& 0.886 $\pm 1.59 \text{e-}3$  
& \textbf{0.905} $\pm 1.22 \text{e-}3$ \\  
\bottomrule
\end{tabular}
\label{tab:accuracy_dropoutDE}
\end{table}

\newpage

\section{Additional experimental results related to section~\ref{subsec:priors}}

\begin{figure}[h!]
    \centering
    \subfloat[1st connection ML ]{\includegraphics[width=0.48\linewidth]{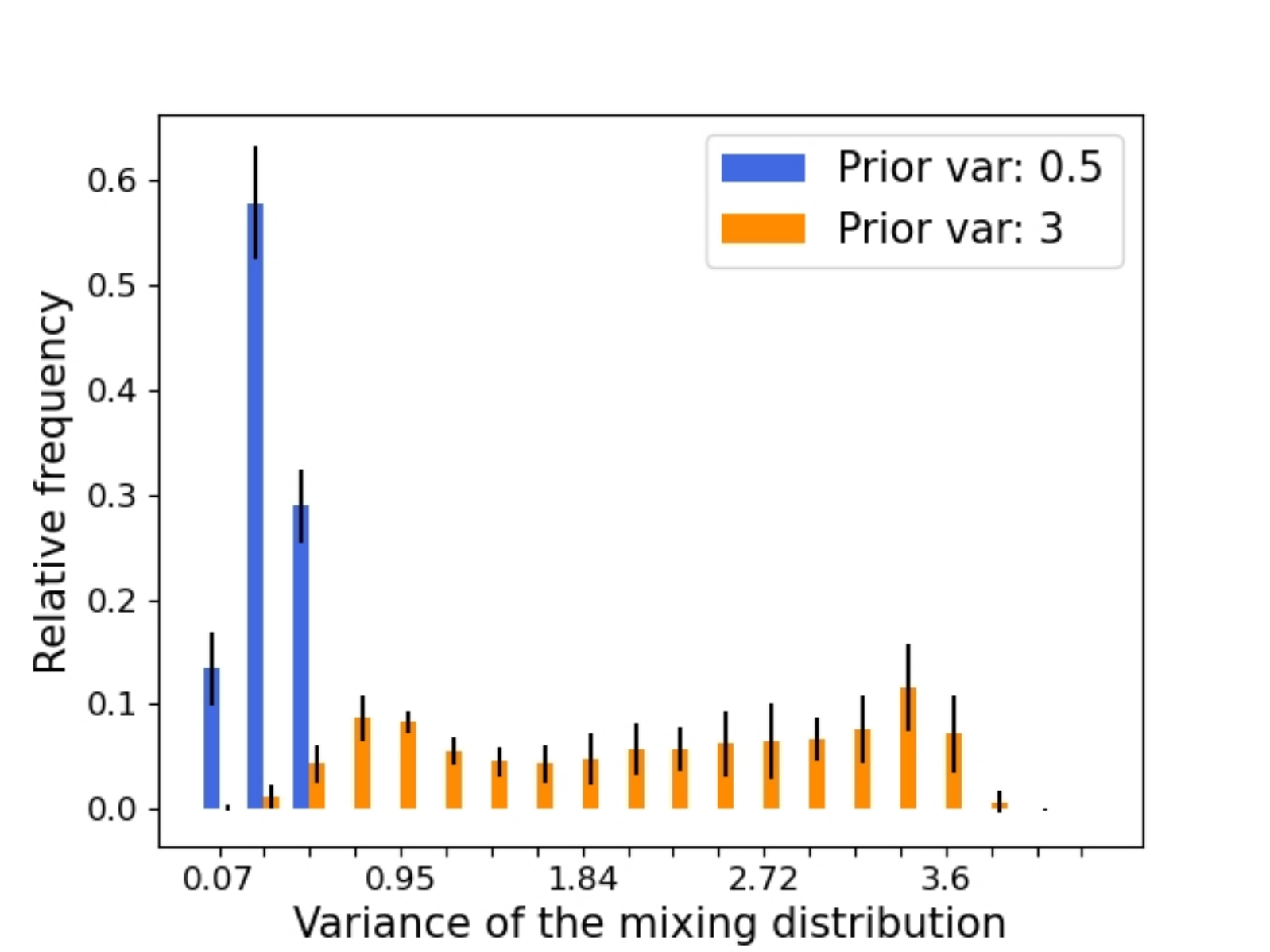}}
    \subfloat[1st connection VI]{\includegraphics[width=0.48\linewidth]{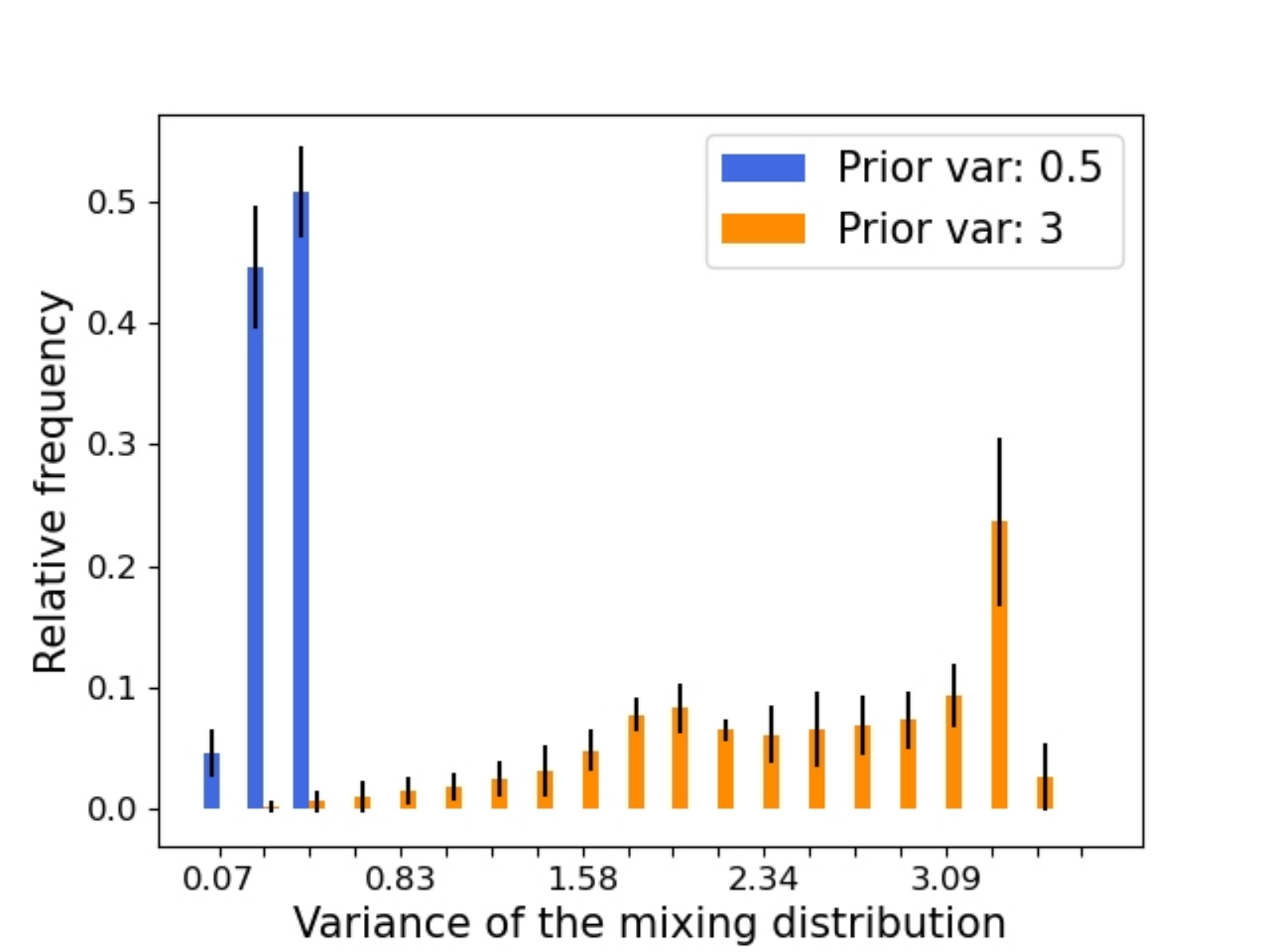}}
    \qquad
    \subfloat[3rd connection ML]{\includegraphics[width=0.48\linewidth]{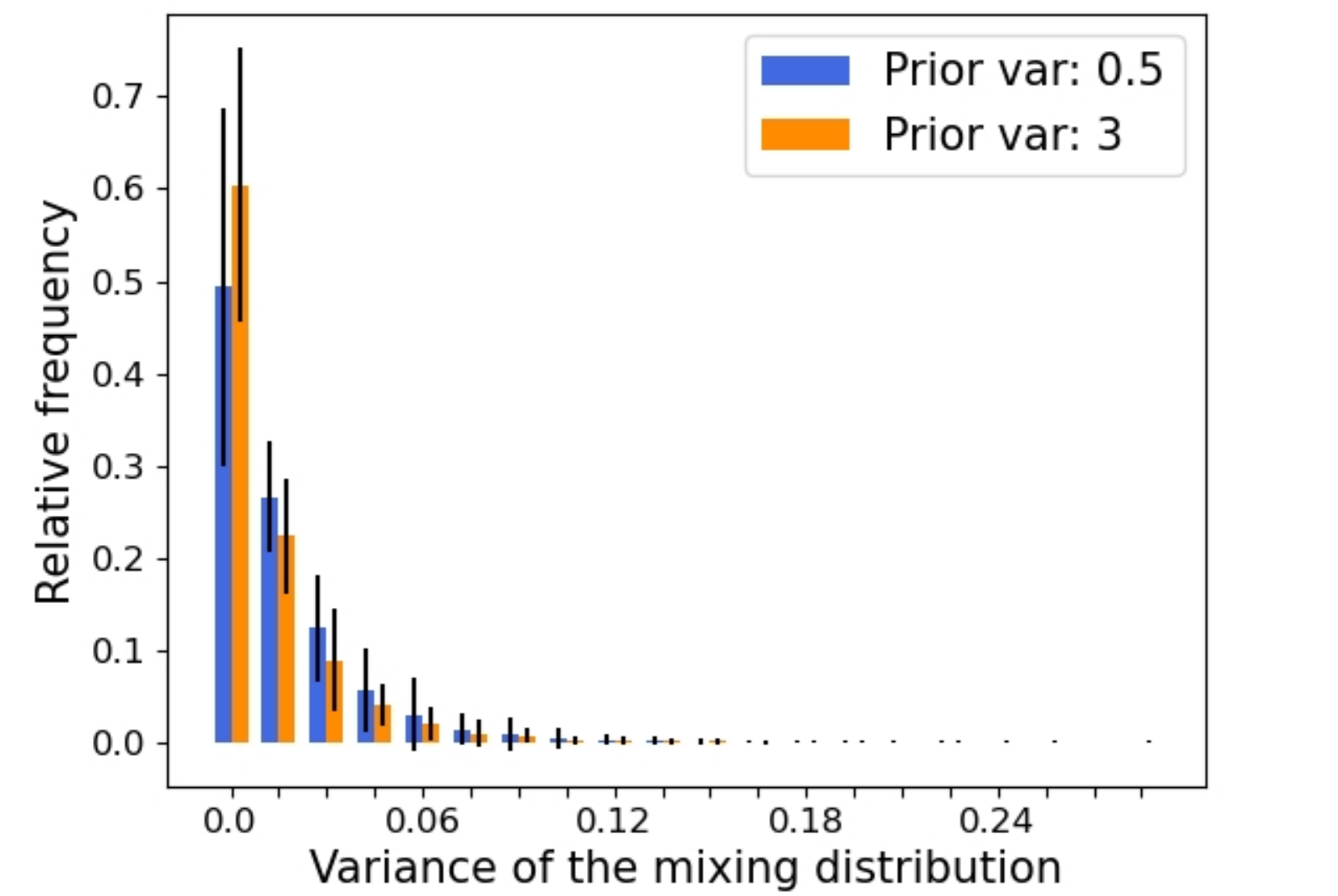}}
    \subfloat[3rd connection VI ]{\includegraphics[width=0.48\linewidth]{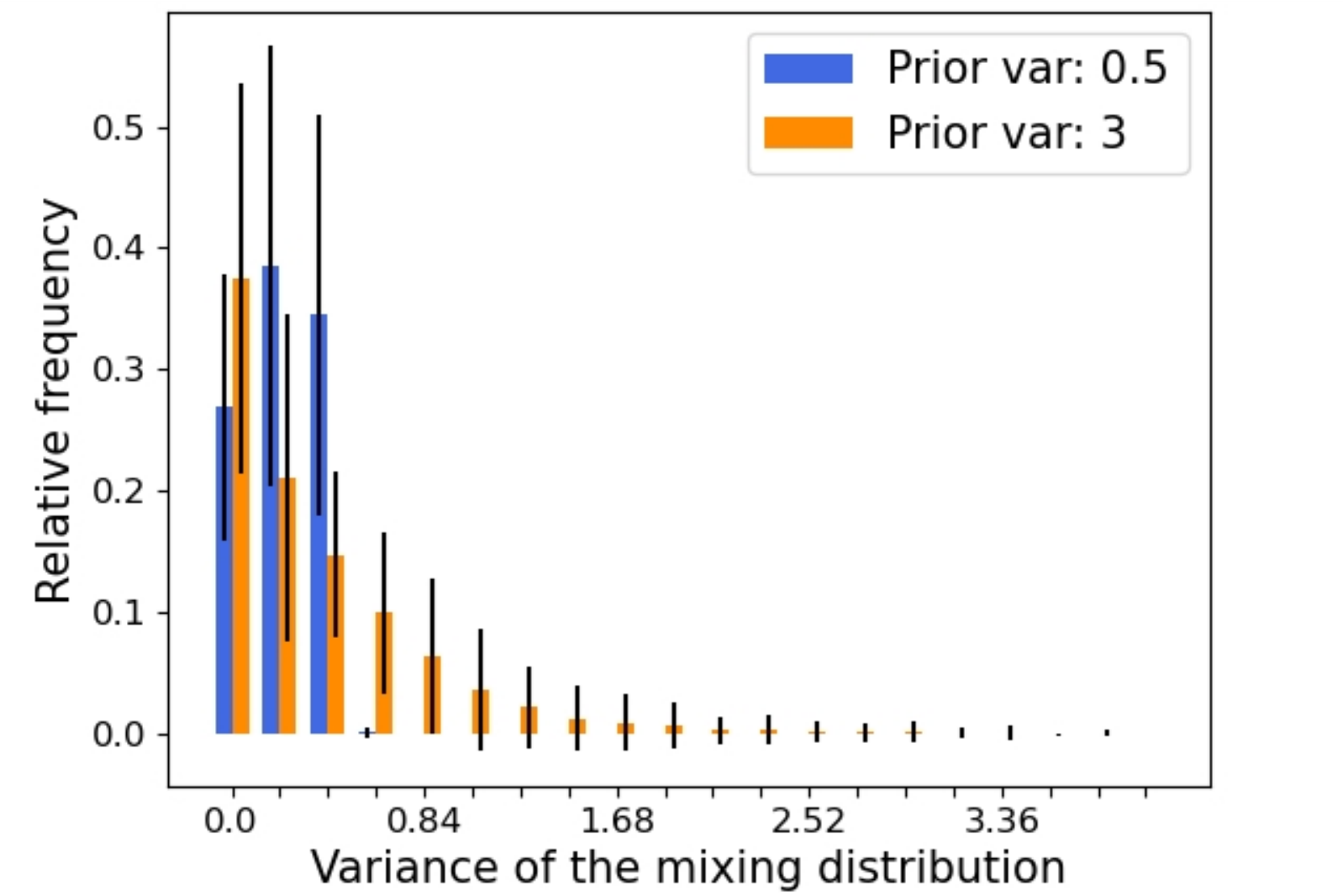}}
    \caption{Variance of the mixing distribution for MNIST trained mixtures. First row: 
    variance of the weights between input and hidden neurons of the first hidden layer
    second row: 
    variance of the weights between the hidden neurons of the second hidden layer and the output neurons.
    }
    \label{fig:app_prior_variance}
\end{figure}
\newpage 
\begin{figure}[h!]
    \centering
    \subfloat[ML MNIST]{\includegraphics[width=0.48\linewidth]{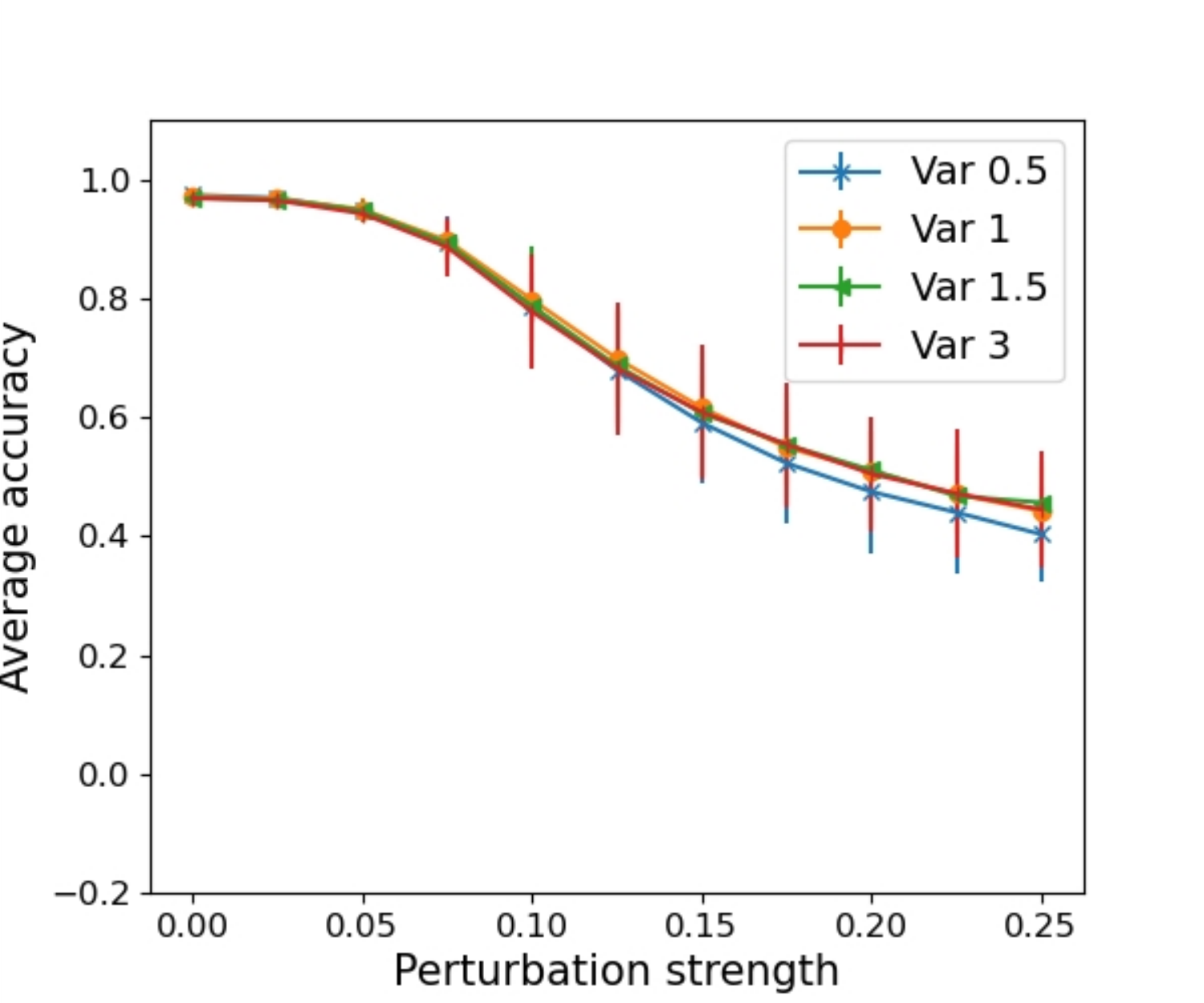}}
    \subfloat[ VI MNIST]{\includegraphics[width=0.48\linewidth]{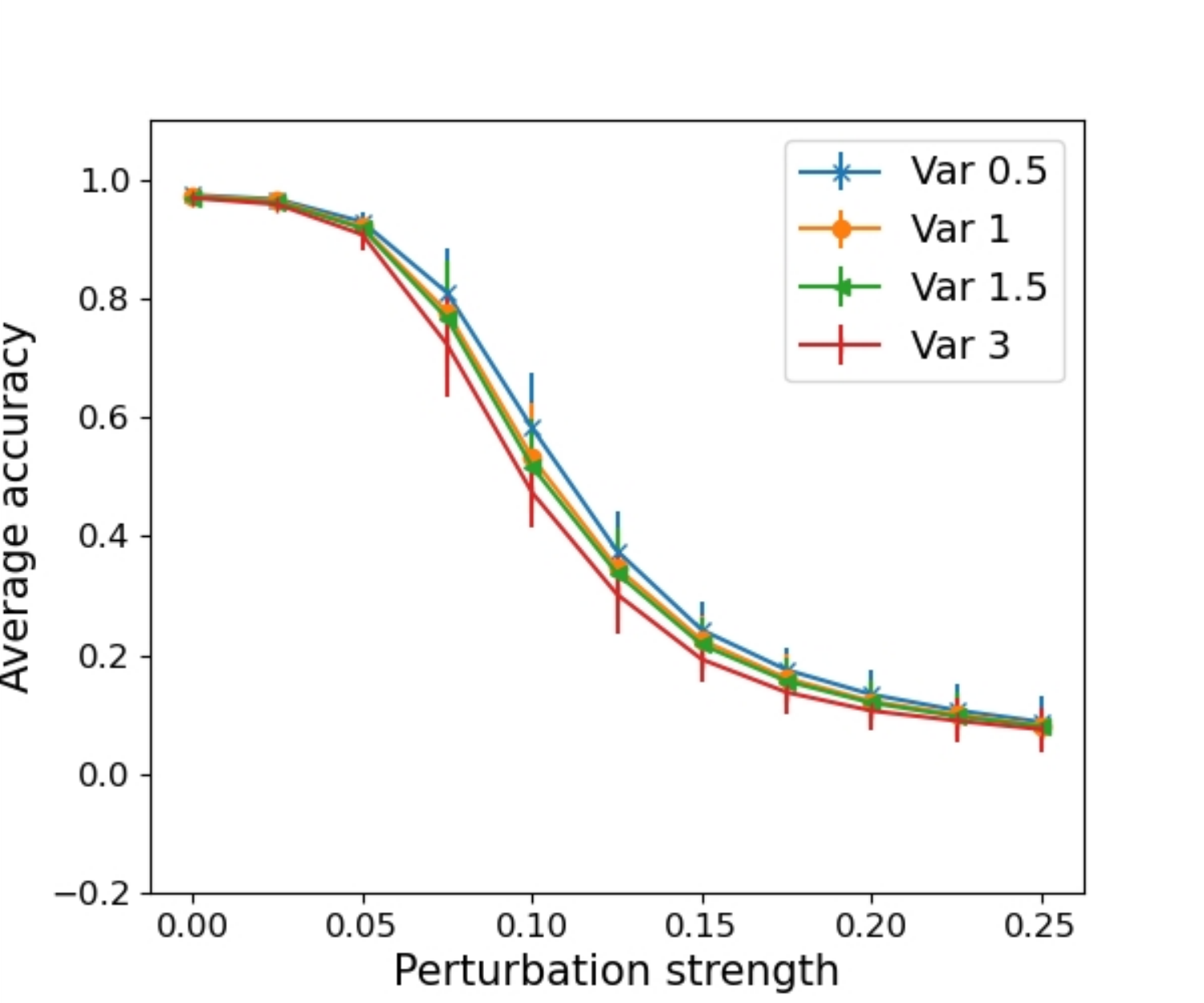}}
    \qquad
    \subfloat[ ML FMNIST ]{\includegraphics[width=0.48\linewidth]{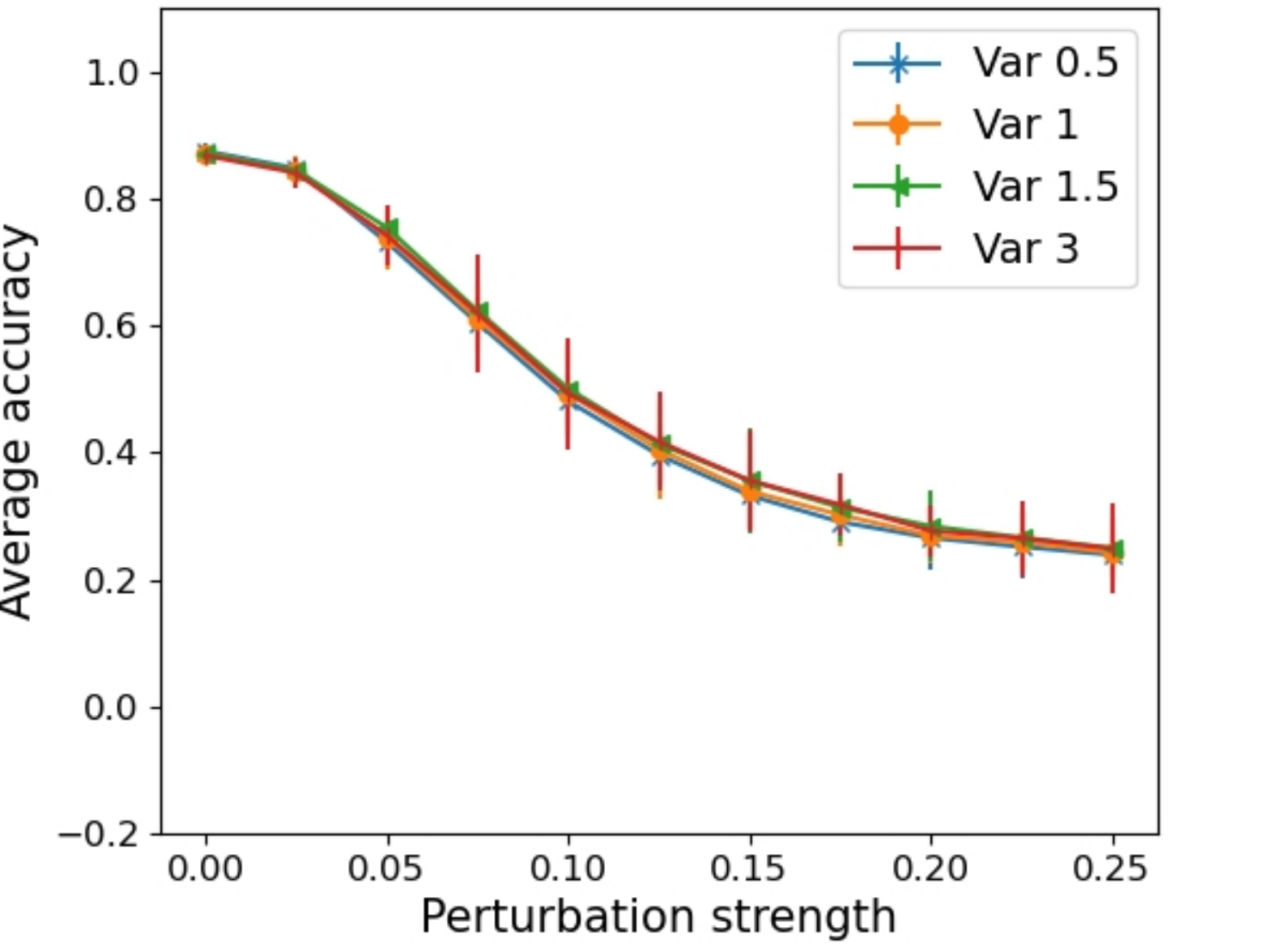}}
    \subfloat[ VI FMNIST ]{\includegraphics[width=0.48\linewidth]{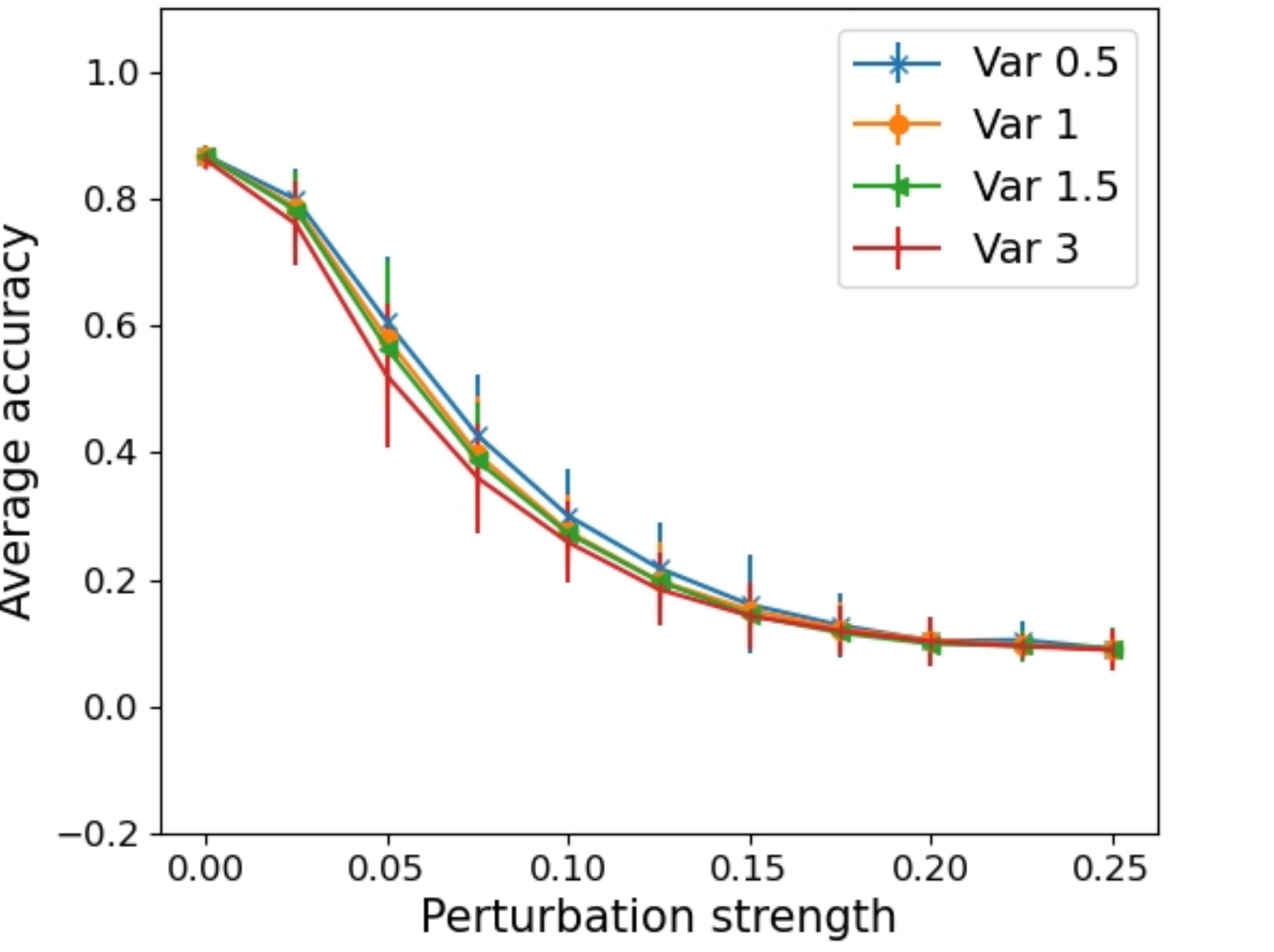}}
    \caption{  Average classification accuracy over 1,000 adversarial examples (3-fold standard deviation indicated by error bars) in dependence of the perturbation strength, for ML- and VI- based models with varying prior variances, top: MNIST, bottom: FMNIST.
    }
    \label{fig:app_prior_accuracy}
    \vspace{-0.5cm}
\end{figure}

\newpage 

\section{Additional experimental results related to section~\ref{subsec:test}}
\begin{figure}[h!]
    \centering
    \subfloat[MNIST - ML ]{\includegraphics[width=0.5\linewidth]{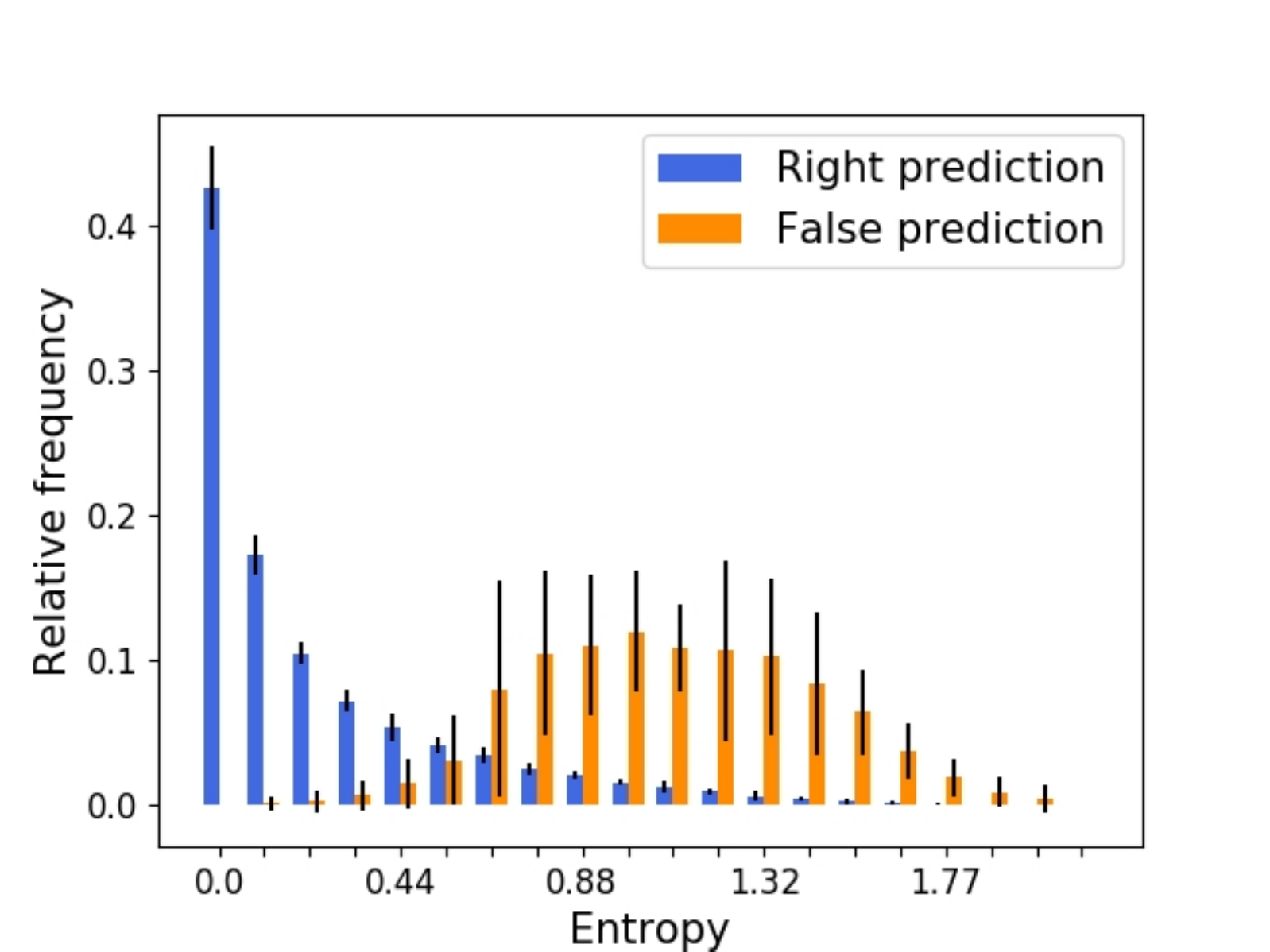}}
    \subfloat[MNIST - VI]{\includegraphics[width=0.5\linewidth]{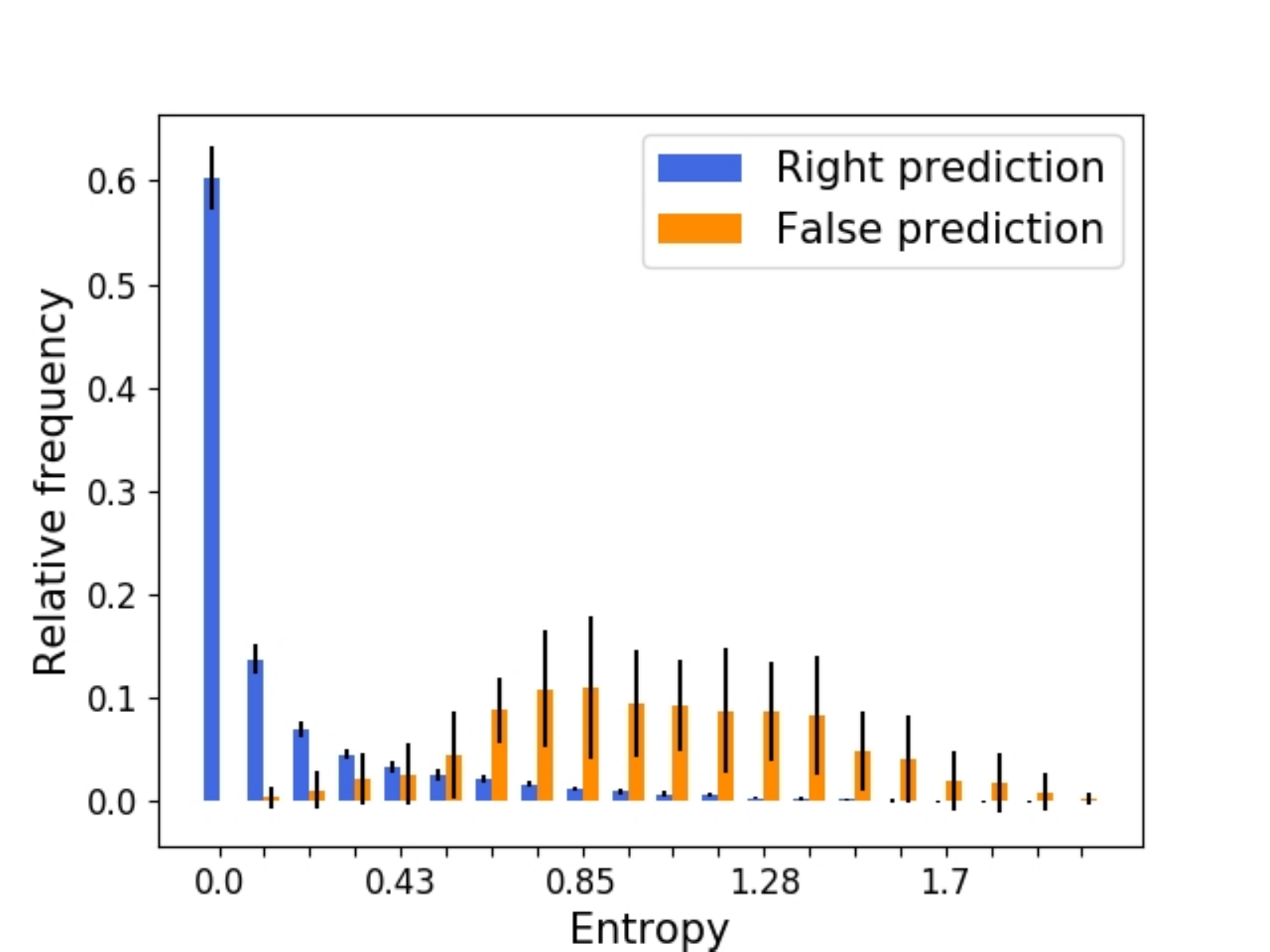}}
    \qquad
    \subfloat[FMNIST - ML ]{\includegraphics[width=0.5\linewidth]{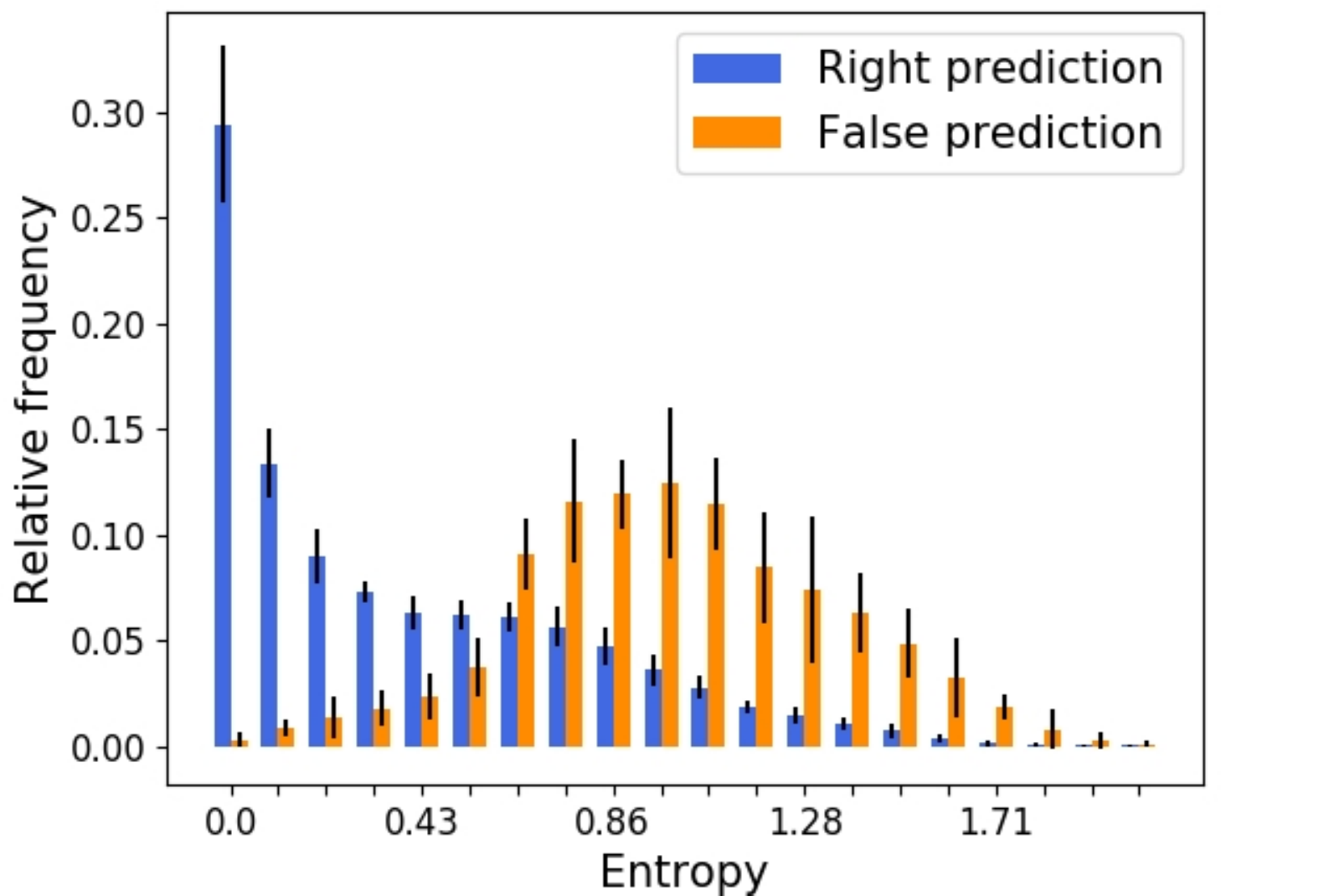}}
    \subfloat[FMNIST - VI]{\includegraphics[width=0.5\linewidth]{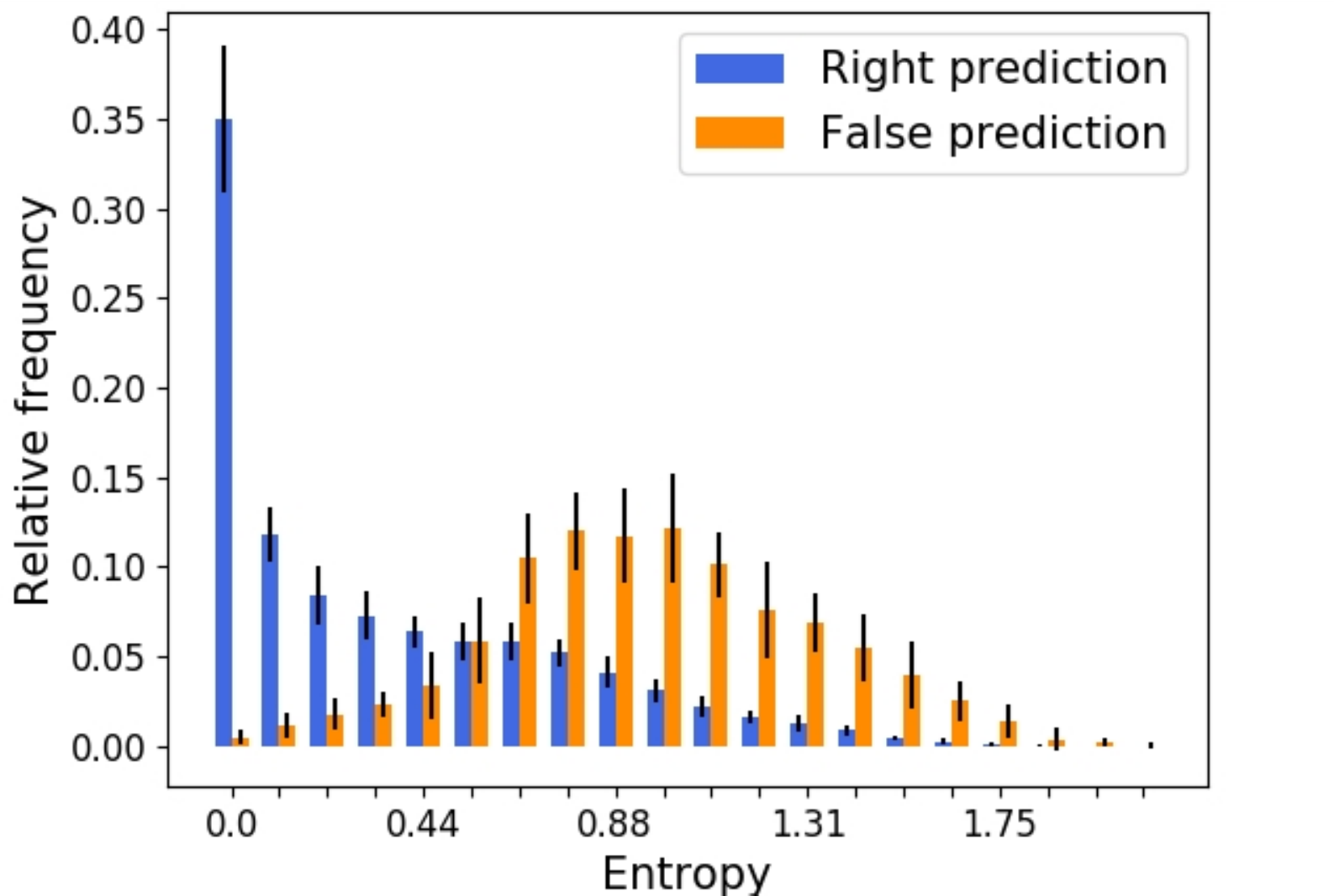}}
   \caption{Entropy of the test set predictions for ML and VI on MNIST and FMNIST divided into correctly and wrongly classified samples. Error bars indicate the 3-fold standard deviation over 10 trials.}
   \label{fig:test_entropy}
\end{figure}

\section{Extension of section~\ref{subsec:detecting_adv_ex}}

In section~\ref{subsec:detecting_adv_ex} we have ignored the fact that the ML objective has a higher accuracy even on attacked examples as we have seen in section~\ref{subsec:stronger_attacks}. Therefore, 
we investigated AUROC scores when only considering the correctly classified test data points as the true class and the wrongly classified adversarial examples (e.g. successful attacks) as the negative class. This leads to unbalanced classes (see table~\ref{tab:numbers} for the  amount of samples per class for one exemplary trial of the experiment).
We account for this by limiting the number of samples in both classes to the minimum over the correctly classified test samples and the number of successful attacks, respectively, 
to calculate the averaged AUROC scores in table~\ref{tab:auroc_adv_correctvssucc}.
These are generally higher than those shown in table~\ref{tab:auroc_adv}, but are in line with out previous findings.

\begin{table}[h!]
\caption{Number of correctly classified test samples and successful attacks (SAdv) for one exemplary trial.}
\centering
\begin{tabular}{l |l l l l } 
\toprule
    &MNIST &&FMNIST & \\
    & \#Test  & \#SAdv & \#Test  & \#SAdv\\ 
\midrule 
VI  & 9,732  $\enspace$ & \textbf{9,042} $\enspace$ & \textbf{8,557} $\enspace$ & 9,092 $\enspace$ \\
ML \ & 9,738  & \textbf{4,818} &  8,580& \textbf{7,477} \\
\bottomrule
\end{tabular}
\label{tab:numbers}
\end{table}

\begin{table}[h!]
\caption{AUROC scores for distinguishing between correctly classified test samples and successful adversarial attacks.}
\centering
\begin{tabular}{l |l l| l l } 
\toprule
    &MNIST && FMNIST & \\
    & ML  & VI & ML  & VI\\ 
\midrule 
Predictive variance $\enspace$  
&0.783 $\pm 5.56 \text{e-2}  \enspace$ 
& 0.623 $\pm 3.02 \text{e-2} \enspace$ 
& 0.551 $\pm 5.44 \text{e-2} \enspace$ 
& 0.367 $\pm 4.40 \text{e-2} \enspace$\\
Entropy \ & \textbf{0.799}$\pm 6.19\text{e-2}  $
&  0.699 $\pm 3.80 \text{e-2}$ 
&  \textbf{0.588}$\pm 6.02 \text{e-2}$ 
& 0.373 $\pm 4.68 \text{e-2}$\\
\bottomrule
\end{tabular}
\label{tab:auroc_adv_correctvssucc}
\end{table}

\end{document}